\documentclass[conference, final]{IEEEtran}
\pdfoutput=1
\usepackage{cite}
\usepackage{algorithmic}
\usepackage{graphicx}
\usepackage{textcomp}
\usepackage{xcolor}
\usepackage{caption}
\usepackage{hyperref}
\usepackage{subfigure}
\usepackage{tikz}
\usepackage{textcomp}
\usepackage{lipsum}
\usepackage{amsmath}
\usepackage{amssymb}
\usepackage{amsfonts}
\usepackage{amsthm}
\usepackage{mathtools}

\DeclarePairedDelimiter{\norm}{\lVert}{\rVert}

\newcommand{\Reals}{\mathbb{R}}

\newcommand{\InputSpace}{\mathcal{I}}
\newcommand{\OutputSpace}{\mathcal{Y}}

\newcommand{\StateSpace}{\mathcal{S}}
\newcommand{\ActionSpace}{\mathcal{A}}

\DeclareMathOperator{\sgn}{sgn}

\DeclareMathOperator*{\argmax}{arg\,max}
\newcommand{\Loss}{\mathcal{L}}

\newcommand{\QFunction}{\mathcal{Q}}
\newcommand{\RewardFunction}{\mathcal{R}}
\newcommand{\Policy}{\pi}
\newcommand{\Discount}{\gamma}
\newcommand{\Bottom}{\perp}

\newcommand{\LatentVector}{\mathbf{h}}

\graphicspath{{./images/}}

\newcommand{\Orig}{m}
\newcommand{\Counter}{m^{\prime}}
\newcommand{\Model}{\varphi}
\newcommand{\EsolPrediction}{s_{\Orig}}
\newcommand{\EsolCounterPrediction}{s_{\Counter}}
\newcommand{\Tox}{TOX21}
\newcommand{\Esol}{ESOL}

\newcommand\copyrightnotice{%
\begin{tikzpicture}[remember picture,overlay]
\node[anchor=north,yshift=-5pt] at (current page.north) {\fbox{\parbox{\dimexpr\textwidth-\fboxsep-\fboxrule\relax}{\footnotesize \textcopyright 2021 IEEE. Personal use of this material is permitted.
  Permission from IEEE must be obtained for all other uses, in any current or future
  media, including reprinting/republishing this material for advertising or promotional
  purposes, creating new collective works, for resale or redistribution to servers or
  lists, or reuse of any copyrighted component of this work in other works.\\
  \emph{To appear in the Proceedings of the 2021 International Joint Conference on Neural Networks (IJCNN 2021)}
  }}};
\end{tikzpicture}%
}

\begin{document}

\title{MEG: Generating Molecular Counterfactual Explanations for Deep Graph Networks}

\author{\IEEEauthorblockN{Danilo Numeroso, Davide Bacciu}
\IEEEauthorblockA{\textit{Department of Computer Science} \\
\textit{University of Pisa}\\
Pisa, Italy \\
danilo.numeroso@phd.unipi.it, bacciu@di.unipi.it}
}
\maketitle
\copyrightnotice{}

\begin{abstract}
Explainable AI (XAI) is a research area whose objective is to increase trustworthiness and to enlighten the hidden mechanism of opaque machine learning techniques. This becomes increasingly important in case such models are applied to the chemistry domain, for its potential impact on humans' health, e.g, toxicity analysis in pharmacology.
In this paper, we present a novel approach to tackle explainability of deep graph networks in the context of molecule property prediction t asks, named MEG (Molecular Explanation Generator). We generate informative counterfactual explanations for a specific prediction under the form of (valid) compounds with high structural similarity and different predicted properties. 
Given a trained DGN, we train a reinforcement learning based generator to output counterfactual explanations. At each step, MEG feeds the current candidate counterfactual into the DGN, collects the prediction and uses it to reward the RL agent to guide the exploration. Furthermore, we restrict the action space of the agent in order to only keep actions that maintain the molecule in a valid state.
We discuss the results showing how the model can convey non-ML experts with key insights into the learning model focus in the neighbourhood of a molecule.
\end{abstract}

\section{Introduction}
The ever-growing predictive capabilities of deep learning models have come at the price of increasing complexity in such models. This contributes to the lack of accountability and transparency of the decision-making which intrinsically functions as a black-box. Explainability of AI predictions is extremely important, especially in AI applied to life sciences and their impact on human lives and health \cite{BacciuBLMOV19}. Chemistry is the life science that
has seen a surge of related work in the deep learning field, covering health-related tasks such as drug-design, drug-discovery, toxicological analysis and molecule-property prediction in general.
The application of deep neural networks in predicting functional
and structural properties of chemical compounds is a research topic with long-standing roots \cite{micheli07}. Deep Graph Networks (DGNs) \cite{zhou2018graph,DBLP:journals/nn/BacciuEMP20} have presently arisen as state-of-the-art for learning effective vectorial molecule representations.
As DGNs become able to solve increasingly complex tasks \cite{pmlr-v70-gilmer17a}, the need of reliable explainability models emerges.
The scarce intelligibility of such models and of the internal representation they develop can, in fact, act as a show-stopper for their consolidation, e.g. to predict safety-critical molecule properties, especially when considering well known issues of opacity in DGN assessment \cite{Errica2020A}.

We present and discuss the most relevant methods regarding explainable deep learning and explainability for graphs in \autoref{sec:rel}.

This paper fits into this latter pioneering field of research by taking a novel angle to the problem, targeting the generation of interpretable counterfactual
explanations for the primary use of the experts of the molecular domain.
While some work on the generation of human-readable explanations through neural networks
does exist \cite{agrawal2015vqa,zhao2020fast} and has been introduced in the context
of DGNs \cite{Yuan2020}, to the best of our knowledge there exists no prior work targeting
counterfactual explanations for graphs.

Hence, we propose MEG (Molecular Counterfactual Generator), a model-agnostic method based on a reinforcement learning \cite{sutton1998reinforcement} explanatory agent. We build our approach upon the assumption that a domain expert would be interested in understanding the model prediction for a specific molecule based on differential case-based reasoning against counterfactuals, i.e. similar structures
which the model being explained considers radically different with respect to the predicted property. Such counterfactual molecules should allow the expert to understand if the structure-to-function mapping learnt by the model is coherent with the consolidated domain knowledge, at least for what pertains a tight neighbourhood around the molecule under study. Our approach is specifically thought for molecular applications and the RL agent leverages domain knowledge to constrain the generated explanations to be valid molecules.

We validate our approach on DGNs tackling the prediction of
different molecule property prediction tasks, namely binary classification on molecule toxicity and regression on solubility
of chemical compounds. First, we run a model selection to pick the best performing model on an held-out validation set. Then, we feed the selected DGN with molecules sampled from an external test set and run a qualitative analysis on its predictions to assess the model behaviour in a plausible operative scenario.

\section{Related Work}\label{sec:rel}

\subsection{Deep Graph Networks}
Starting with their introduction by Micheli \cite{micheli09}, deep graph networks have rapidly become the state-of-the-art in graph-based tasks. Deep Graph Networks are an evolution of the classical deep neural networks that are able to work and learn patterns directly from graphs. 
In graph theory, a graph is defined as a pair $G = (V, E)$ where $V$ is a set of entities (nodes) composing the graph and $E \subseteq V \times V$ represents the relations, i.e. link between nodes, called edges. As far as DGNs are concerned, the plain definition is enriched with node and edge features that are stored in the node-feature matrix $X \in \Reals^{n \times d_v}$ and edge-feature matrix $A \in \Reals^{m \times d_e}$, respectively. Therefore, DGNs perform operations on graphs $G = (V, E, X, A)$.

DGNs learn by computing information for each node and diffusing it across all the graph. This mechanism is known as context diffusion. At each step, DGNs aggregate information from the node and its neighbours and use it to update the node representation. 
In general, DGN models can differ on how the context diffusion mechanism is implemented in practice \cite{kipf2017semisupervised, hamilton2017inductive, bacciujmlr2020}.
In its general form, the neighbourhood aggregation follows a 2-step message passing algorithm: (i) each node in the graph computes a message as a function of its current state $\LatentVector^\ell_v$ and (possibly) edge information. Then, the message is sent to all the neighbours; (ii) each node computes its next state $\LatentVector^{\ell+1}_v$ by aggregating the messages that come from its neighbours.
Formally, the neighbourhood aggregation process can be formalised as:
\begin{equation}
    \LatentVector^{\ell+1}_v = \phi^{\ell+1} \big (
        \LatentVector^{\ell+1}_v,
        \Psi(\{ \psi^{\ell +1}(\LatentVector^{\ell}_u) \mid u \in \mathcal{N}_v\})
    \big )
    \label{eq:node-aggr}
\end{equation}
where $\phi$ and $\psi$ are transformations of the input data and
$\mathcal{N}_v$ represents the neighbourhood of node $v$.
$\Psi$ is a permutation invariant function, e.g. sum, which is required to discard the effect of arbitrary ordering of the set of neighbours. In other words, the output is not influenced by the order in which we evaluate and aggregate neighbours messages. 
$\LatentVector^{\ell}_v$ represents the node state (or node representation) at layer $\ell$. For $\ell=0$, each node $v$ is initialised with the associated feature vector $x_v \in X$, i.e. $\LatentVector^0_v = x_v$.

\subsection{Explainable AI}
The explainability of AI systems is a research topic that
aims to explain why and how black-box models, e.g. neural networks, arrive at a specific decision.
In general, explainability approaches can be distinguished in {\it global} vs {\it local}. Briefly, a global explanation method aims to provide insights into the model behaviour as a whole, which is usually difficult to obtain as it requires to find a concise description of the model that can be used to predict the future behaviour as well. Conversely, a local explainer relaxes this assumption by explaining how the model behaves for a specific sample (or a sufficiently wide neighbourhood of the input).
Any explainability method can be further categorised as {\it model-specific} or {\it model-agnostic}, based on whether the method can explain only certain model architectures, i.e. the former, or any model by treating it as a black-box, i.e. the latter.

The most advanced area within this field of research regards the explainability for images and vectorial data. 
One class of methods are based on sensitivity analysis and falls into the local model-specific category. 
Symonian, Vedaldi and Zisserman \cite{simonyan2013deep} targets the explanation for an image $I$ of class $c$ in Convolutional Neural Networks (CNNs) by taking the derivative of the class score function $S$ with respect to the input image $\frac{\delta S}{\delta I}$. This derivative is then used to build a salience map indicating the importance of each 
input feature, i.e. pixels.
CAM \cite{zhou2015learning} is an extension of the previous method that aims at finding class-specific feature maps in the latent space, by taking the previous derivative with respect to the penultimate layer of the CNNs. In CAM, the penultimate layer is forced to be convolutional, making the whole method model-specific.
LIME \cite{lime} takes a different approach, tackling the
explanations of complex models by approximating them locally with simpler interpretable models. Such models approximate complex models in a tight locality of a given input $x_0$, obtained by perturbing $x_0$. As LIME makes no assumptions on the architecture of the model being explained, it falls into the category of model-agnostic methods.

Other approaches within the input perturbation methods include counterfactual explanations. A counterfactual explanation is defined as, starting from a given input, a small perturbation of the starting input configuration which produces a desired prediction (possibly radically different). Furthermore, a good counterfactual must confuse the network being explained while aligning with the underlying data distribution. In other words, it should not present an unlike configuration of features which can easily fool the model.
Finding counterfactual explanations can then be defined as an optimisation problem \cite{cf}. Van Looveren and Klaise \cite{vanlooveren2020interpretable} apply counterfactual discovery in classification tasks by using class prototype to align the produced counterfactuals to the data distribution. Different ways to produce counterfactuals include an adversarial term in the objective function \cite{zhao2020fast}.

Explainability in the field of deep learning for graphs is, however, much less explored.
Although some early-stage effort has been made to generalise established explainability techniques to DGNs \cite{baldassarre2019explainability, pope2019explainability}, these methods turned out to not be as effective in the graph scenario.
In this respect, attention must be concentrated towards the development of interpretability techniques specifically tailored to DGNs. While some DGN shows potential for interpretability {\it by-design} thanks to its probabilistic formulation \cite{bacciujmlr2020}, the majority of works in literature take a neural-based approach which requires the use of an {\it external} model explainer.
GNNExplainer \cite{ying2019gnnexplainer} is the front-runner of the model-agnostic
methods providing local explanations to neural DGNs in terms of the
sub-graph and node features of the input structure which maximally
contribute to the prediction. RelEx \cite{zhang2020relex} extends
GNNExplainer to surpass the need of accessing the model gradient to
learn explanations. GraphLIME \cite{huang2020graphlime} attempts to
create locally interpretable models for node-level predictions, with
application limited to single network data.
CoGE \cite{faber2020contrastive} introduced contrastive explanations
restricted to graph classification problems. Specifically, given a graph
along with its prediction, CoGE partitions the training data into the set
of graphs having the same label as of the input graph and those with different
labels. Thus, it selects a number of graphs from the two sets that are most
similar to the input graph and optimize the Optimal Transport (OT)
\cite{nikolentzos2017matching} to obtain the nodes (and edges) that are responsible
for the prediction.
Finally, as far as life sciences are concerned, the current literature seems lacking in interpretability methods specifically thought for this field.
CellBox \cite{yuan2020interpretable} is one of the few works on the topic. The authors couple the application of a machine learning framework with explicit mathematical models for modeling cellular response to perturbation. The aim is to learn the interactions between cellular components in a supervised setting, combining the learnt features with an ODE solver to enhance the interpretability of the whole model. Therefore, this work aims to achieve interpretability by-design rather than requiring the use of an external explainer.
More in general, some of the general interpretability approaches can find room for application in life sciences too \cite{azodi}.  

\section{Molecular Explanation Generator (MEG)}\label{sect:model}
\begin{figure*}
    \centering \includegraphics[width=0.8\linewidth]{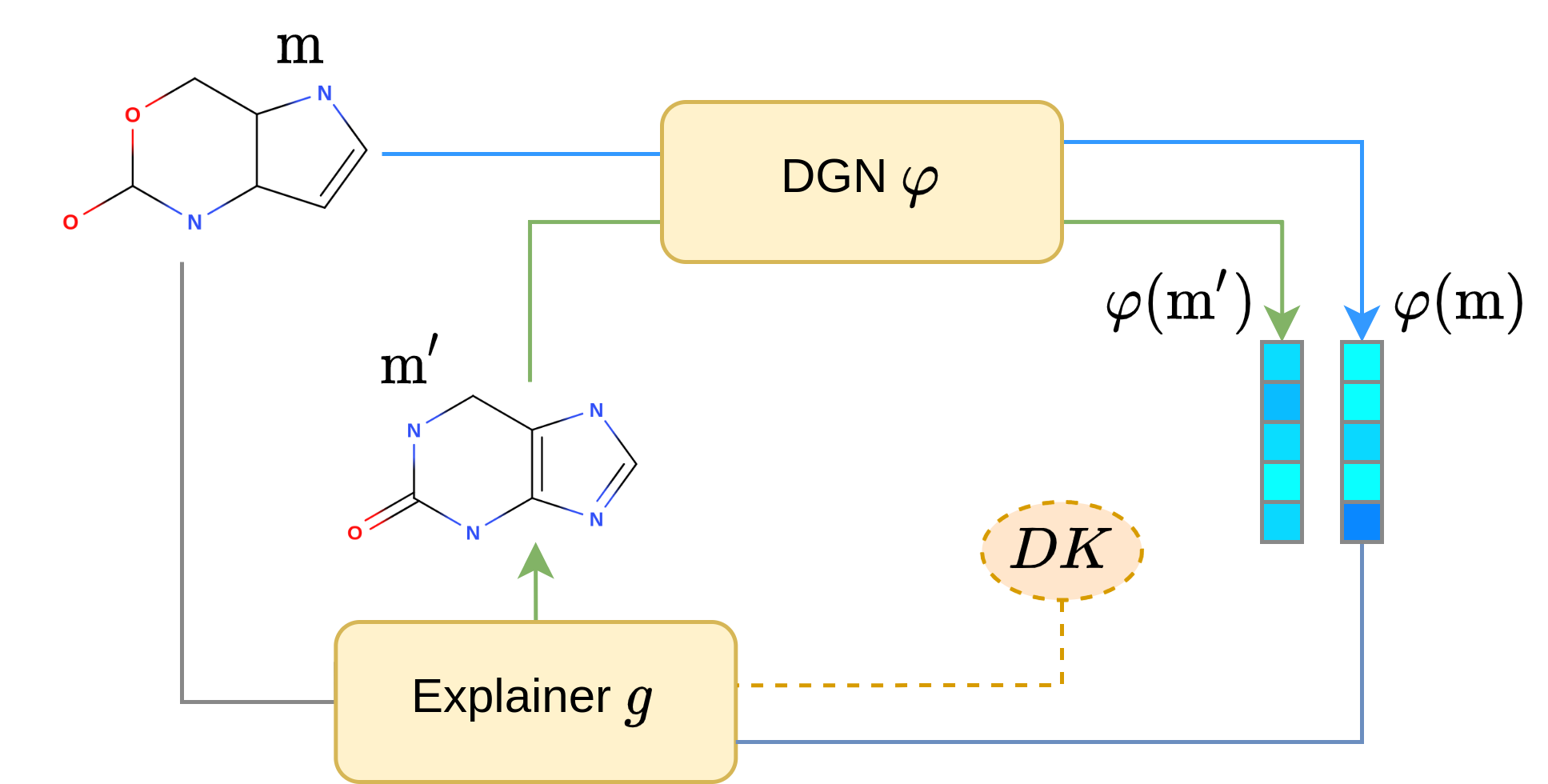}
    \caption{Architecture of the Molecular Explanation Generator (MEG): DGN $\Model$ is a trained molecule property predictor, whereas the Explainer $g$ is a generative agent producing counterfactuals, constrained by prior domain knowledge $DK$.}
    \label{fig:architecture}
\end{figure*}

\subsection{Method Description}
The overall architecture of MEG is presented in \autoref{fig:architecture}.
We aim to explain the depicted DGN $\Model: \InputSpace \rightarrow \OutputSpace$ that is fit to solve a molecule property prediction task. 
$\InputSpace$ represents the space of (labelled) molecule structures and $\OutputSpace$ is the task-dependent output space.
We assume the DGN to: (i) build vectorial graph representations out of the graph topology and associated information; (ii) grant access to these internal representations.

The explainer $g$ takes care of generating counterfactual explanations under the form of molecular graphs. The problem of learning to generate generic graph structures is a complex one \cite{BACCIU2020177}, although some effective solutions exist that are specialized on molecule generation \cite{PoddaBM20}. The generative task tackled by the explainer is, luckily, less general and it is well suited to be implemented by a reinforcement learning based approach. Intuitively, given the sample $\Orig$, the explainer $g$ is trained to identify minimal modifications to $\Orig$ that maximally change the outcome of the model prediction, obtaining eventually the counterfactual $\Counter$ in the figure. Through these changes, one may evaluate the model robustness when it comes to predicting out-of-distribution inputs or finding out critical difficulties, guaranteeing a more thorough understanding of the predictive model itself.
In graph structured data, counterfactuals can be obtained by either perturbing
the node (or edge) features or the graph topology. The former boils down to perturbation of one-dimensional feature vectors, which is a problem that has been 
widely explored in the literature and relate to the explainability in a non-relational case \cite{lime}. Thus, in this work we focus on finding explanations for a DGN prediction by exclusively alter the graph structure.
% in a discrete manner.
MEG directly operates on the input molecule topology by means of discrete graph alteration steps.

Molecular counterfactuals ought to satisfy three properties: (i) they need to resemble the molecule under study; (ii)
predicted properties on counterfactuals must differ substantially from
those predicted on the input one; (iii) molecular counterfactuals need
to be in compliance with chemical constraints. To this end, the agent
$g$ receives information about an input molecule $\Orig$ and its
associated prediction score $\Model(\Orig)$, and generates a
molecular counterfactual $\Counter$, leveraging prior domain knowledge
to ensure validity of the generated sample. 
Counterfactual generation is formalised as a maximisation problem in which, given a target molecule $\Orig$ with prediction $\Model(\Orig)$, the generator $g$
is trained to optimize:
\begin{equation}\label{eq:L}
    \argmax_\theta \Loss \big( \varphi(\Orig), (\varphi \circ g)(\cdot
    \mid \theta) \big) + \mathcal{K} \big [\Orig, g(\cdot \mid \theta)
      \big].
\end{equation}
Where $\theta$ indicates the explainer learnt parameters. 
The composition $(\varphi \circ g)(\cdot|\theta)$ formalizes the
model $\varphi$ counter-predictions, made over the counterfactuals
produced by $g$. 
Given the counterfactual $\Counter = g(\cdot \mid
\theta)$ we rewrite \eqref{eq:L} as
\begin{equation}\label{eq:counterfactual-gen}
\argmax_{m^{'}} \Loss \big( \varphi(\Orig), \varphi(\Counter) \big) +
\mathcal{K} \big [\Orig, \Counter \big]
\end{equation}
where $\Loss$ is a measure of dissimilarity between the
model predictions for the molecule $\Orig$ and its counterfactual $\Counter$, while
$\mathcal{K}$ measures the structural similarity between the molecules $(\Orig,\Counter)$ themselves.
Note that the $\argmax$ operator is applied 
on both $\Loss$ and $\mathcal{K}$ so that the generator can optimise the two properties jointly.

The main use of counterfactual explanations is to provide insights
into the function learnt by the model $\varphi$. In this sense, a set
of counterfactuals for a molecule may be used to: (i) identify changes
to the molecular structure leading to substantial changes in the
properties, enabling domain experts to discriminate whether the model
predictions are well founded; (ii) validate existing interpretability
approaches, by running them on both the original input graph and its
related counterfactual explanations. The main idea behind this latter
point is that a local interpretation method may provide explanations
that work well within a very narrow range of the input, but do not
give a strong suggestion on a wider behaviour. To show evidence and
usefulness of such a differential analysis, in section \ref{sec:experimental-eval}
we use our counterfactuals to assess the quality of explanations given
by GNNExplainer \cite{ying2019gnnexplainer}. Given the undirected
nature of the graphs in our molecular application, we restrict the
original GNNExplainer model to discard the effect of edge orientation
on the explanation.

\subsection{Explanation Generator}
The explanation generator $g$ has access to
the internal representation of the property-prediction model as well
as to its output and uses this information to guide the exploration of
the molecular structure space to seek for the nearest counterfactuals.
Given the non-differentiable nature of the graph alterations and its ease in modelling and handling multi-objective optimization, we model $g$
through a multi-objective RL problem \cite{liu2015morl}. This allows to easily steer towards the
generation of counterfactuals optimizing several properties at a time
\cite{sanchez2017,popova2018}. 

Formally. the problem takes the form of a Markov Decision Process (MDP) ($\StateSpace$, $\ActionSpace$, $\QFunction$, $\Policy$, $\RewardFunction$, $\Discount$). The set of states $\StateSpace$ includes all the possible molecules
that can be generated by MEG. In our framework, $\Orig$ is used to bootstrap the generative process in $g$, i.e, $s_0 = m$ which operates on the current candidate counterfactual with graph editing operations under domain knowledge constraints. 

The action space $\ActionSpace$ requires particular attention to ensure
that all the generated counterfactuals complies to natural chemical rules.
Thus, domain knowledge must be imbued into $\ActionSpace$.
To achieve validity of the counterfactuals, we base the implementation of the generator on the MolDQN \cite{zhou2018optimization} model. The key element is that we borrow the structure of the action space, in which given the current state $s_t$, we obtain $\ActionSpace_t$ as 
union of three different sets of actions: 
\begin{itemize}
    \item $\ActionSpace_a$, made up of all possible atom additions. First,
    we add a new atom $v$ to the atom sets, i.e, $V' = V \cup \{v\}$ and then
    bind it to an atom $u \in V$. If the resulting molecule does not violate
    any chemical constraints, the action is retained. Otherwise it is discarded.
    
    \item $\ActionSpace^+_b$, representing the set of bond additions. We scan the set of atoms $V$ pair-wisely and either add a new bond or increase the bond order in case the two current atoms are already linked, as follows: 
    \begin{enumerate}
        \item No bond $\longrightarrow$ \{Single, Double, Triple\} bond
        \item Single bond $\longrightarrow$ \{Double, Triple\} bond.  
        \item Double bond $\longrightarrow$ \{Triple\} bond.
    \end{enumerate}
    
    \item $\ActionSpace^-_b$, representing the set of bond removals which is obtained by reversely applying the bond addition rules.
\end{itemize}
Therefore, they are all combined to construct the action space $\ActionSpace_t = \ActionSpace_a \cup \ActionSpace^+_b \cup \ActionSpace^-_b \cup \{\Bottom\}$, where $\Bottom$ is a special action indicating to leave the molecule as is. Eventually, $\ActionSpace_t$ is constrained to discard actions that would lead the molecule in an invalid state, e.g. violation of valence rules.

The multi-objective reward function $\RewardFunction$ exploits the prediction from $\Model$ so as to notify the agent of its current performance, emitting a scalar reward. The term $\gamma$ is called discount and is incorporated into $\RewardFunction$ to weight the contribution of distant future reward signals. 
In our design, $\RewardFunction$ binds together a term regulating the change in prediction scores, which is inherently task-dependent, with a second term controlling similarity between the original molecule and its counterfactual, as presented in \eqref{eq:L}. 

We explored three different formulations for the latter term. First, the Tanimoto similarity over the binary Morgan fingerprints \cite{rogers2010}:
\begin{equation}
    \mathcal{T}(a, b) = \frac{a \cdot b}{\norm{a}^2
    + \norm{b}^2 - a \cdot b} .
\end{equation}
As this metric looks for matching molecule fragments, it may be too susceptible to graph alteration actions. To surpass this issue, we alternatively compare the molecule similarity in the DGN latent space. Thus, we leverage the model's own perception of similarity between molecules by directly comparing the internal vectorial molecule representations through cosine similarity: 
\begin{equation}
    cos(\Orig, \Counter) = \frac{\LatentVector_{\Orig} \cdot \LatentVector_{\Counter}}{\norm{\LatentVector_{\Orig}} \norm{\LatentVector_{\Counter}}.
    }
\end{equation}
However, in our empirical settings we found it useful to combine the model perception
with a fragment-based metric such as the tanimoto similarity. For this reason, we derive a third similarity metric which is a convex combination of the tanimoto similarity (structural) and the neural encoding similarity (model perception):
\begin{equation}\label{eq:cc-sim}
    \mathcal{K}(\Orig, \Counter) = \alpha_1 \mathcal{T}(\Orig, \Counter) + \alpha_2 cos(\Orig, \Counter) \mid \sum_{i} \alpha_i = 1 .
\end{equation}

Lastly, given the current state $s_t$ and a sampled action $a_t$, the explainer evaluates the goodness of the pair action-state through the $\QFunction$ function. Since the state space can become combinatorially large, $\QFunction$ is approximated by a double deep-Q-network \cite{hasselt2015deep} composed of four linear layers.
Finally, the policy $\pi$ is a function outputting the best possible action given any state $s \in \StateSpace$. We use either a decaying $\varepsilon$-greedy policy or an $\varepsilon$-greedy policy to balance exploration and exploitation, depending on whether we are targeting the optimisation of a single counterfactual or a group of them.

\subsection{Training the Generator}
When it comes to training the generator, we must specify the task-dependent term $\Loss$ in the reward function \eqref{eq:counterfactual-gen}. First of all, in a multi-objective reward function the reward is returned as a vector of partial scalar rewards $r_i$, with one partial reward for each property we are seeking to optimize. An intuitive way to combine all the partial rewards is by convex combination:
\begin{equation*}
    r^t = \sum_i \alpha_i r^t_i
\end{equation*}
subject to $\sum_i \alpha_i = 1$. The choice of the coefficients is treated as a model hyper-parameter.

The task-dependent term $\Loss$ in \eqref{eq:counterfactual-gen} can be
specialized for classification and regression tasks. As regards
classification, given a set of classes $\mathcal{C}$, a model
$\Model$ emits a probability distribution $\Model (\cdot) =
\boldsymbol{y} = [y_0, ..., y_{|\mathcal{C}|}]$ over the predicted
classes. In this case, given an input-prediction pair $\langle m, c =
\argmax_{c \in \mathcal{C}} \Model(m) \rangle$, the generator is
trained to produce counterfactual explanations $\Counter$ minimising
the prediction score for class $c$, as follows
\begin{equation}
    \argmax_{\Counter} -\alpha y_c + (1 - \alpha) \mathcal{K}
           [\Counter, \Orig]
      \label{eq:tox21-obj1}
\end{equation}
where $\alpha \in [0,1]$.
Hence, the model $\Model$ returns at each step a smooth reward, which
is actually the inverse of the probability of $\Counter$ belonging to
class $c$. Differently, we define the loss objective for a regression task as
\begin{equation}
    \argmax_{\Counter} \alpha \beta \norm{\EsolCounterPrediction - \EsolPrediction}_1 + (1 - \alpha) \mathcal{K}[\Counter, \Orig] \label{eq:esol-obj}
\end{equation}
where $s$ is the regression target and
$s_{\Orig}$ and $s_{\Counter}$ are the predicted values for the original molecule and its counterfactual, respectively.
Lastly, $\beta = \sgn\big(\norm{\EsolCounterPrediction - s}_1 - \norm{\EsolPrediction - s}_1\big)$ prevents the agent from generating molecules whose predicted scores move towards the original target.

\section{Experimental Evaluation} \label{sec:experimental-eval}

\subsection{Datasets and model configuration}
We test MEG on two popular molecule property prediction benchmarks:
\Tox{} \cite{dortmund2016},
addressing toxicity prediction as a binary classification task, and
\Esol{} \cite{wu2017moleculenet}, that is a regressive task on water
solubility of chemical compounds.

Preliminarily, we scanned both datasets to filter non-valid chemical
compounds. We considered structures to be valid molecules if they pass
the RDKit \cite{landrum2006rdkit} sanitization check. In the end, \Tox{}
comprises 1756 samples, equally distributed among the two classes,
while \Esol{} includes 1129 compounds.
We choose a split of 80\%/10\%/10\% for training, validation and test
set in both datasets.

The trained DGN comprises three GraphConv \cite{morris2018weisfeiler}
layers with ReLu activations. The model has been implemented by using
PyTorch Geometric \cite{Fey/Lenssen/2019} and trained with Adam optimizer.

We ran a model selection and select the model that achieved the highest result on the validation set. We optimized the hyper-parameters by grid-search over the learning rate
in \{$5 \times 10^{-3}, 1 \times 10^{-3}, 5 \times 10^{-4}, 1 \times 10^{-4}$\},
batch size in \{20, 60, 120\} and hidden size in \{32, 64, 128, 256\}. The best configuration for both models is reported in \autoref{tab:config}.
\begin{table}
    \centering
    \begin{tabular}{c|c|c|c|c|c|c}
         Dataset & HS & Batch Size & LR & Optimizer & TR & VL  \\
         \hline
         \Tox{} & 256 & 20 & $1 \times 10^{-3}$ & Adam & 94\% & 88\% \\
         \Esol{} & 32 & 20 & $5 \times 10^{-4}$ & Adam & 0.31 & 0.57 \\
    \end{tabular}
    \caption{HS = hidden size, LR = learning rate, TR = training, VL = validation}
    \label{tab:config}
\end{table}

The network builds a layer-wise molecular representation via concatenation of max and mean pooling operations, over the set of node representations. The final neural encoding of the molecule is obtained by sum-pooling of the intermediate representations. This neural encoding is then fed to a three-layer feed-forward network, with hidden sizes of [128, 64, 32] and dropout set to 0.1, to perform the final property prediction step. Then, we feed the model with molecules in the test set on which it
achieved 77\% of accuracy for \Tox{} and 0.57 MSE for \Esol{}, and try to explain the predictions.

During generation, we employed MEG to find 10 counterfactual explanations for each molecule in test, ranked according to the multi-objective score in Section \ref{sect:model}. In order to
comply with the similarity criterion, we set as starting point for the generation the original test molecule and limit the episode length to 1. Furthermore, we remove $\Bottom$ from the list of available actions and use a decaying epsilon greedy policy $\varepsilon_{t+1} = \lambda \varepsilon_t$ with $\lambda = 0.9987$. Finally, we train MEG for 3000 epochs.
\begin{table}
    \centering 
    \begin{tabular}{c c c c c c} 
        Molecule & Target & Prediction & Similarity & Reward\\ 
        \hline
        A0: \autoref{fig:t1} & {\tt NoTox} & {\tt NoTox} (0.99)&
            - & -\\
        A1: \autoref{fig:t1} & - & {\tt Tox} (0.99) & 0.81 &
            0.86\\
        A2: \autoref{fig:t1} & - & {\tt Tox} (0.99) & 0.74 &
            0.85\\
        A3: \autoref{fig:t1} & - & {\tt Tox} (0.99) & 0.74 & 
            0.85\\
        B0: \autoref{fig:t2} & {\tt Tox} & {\tt Tox} (0.99)&
            - & -\\
        B1: \autoref{fig:t2} & - & {\tt NoTox} (0.93) & 0.77 &
            0.78\\
        B2: \autoref{fig:t2} & - & {\tt NoTox} (0.89) & 0.74 &
            0.77\\
        C0: \autoref{fig:e1} & -4.28 & -4.01 & - & -\\ 
        C1: \autoref{fig:e1} & - & -6.11 & 0.63 & 1.37\\
        C2: \autoref{fig:e1} & - & -5.93 & 0.64 & 1.23\\
        C3: \autoref{fig:e1} & - & -5.07 & 0.63 & 1.18\\
        D0: \autoref{fig:e2} & -5.23 & -5.31 & - & -\\
        D1: \autoref{fig:e2} & - & -6.95 & 0.71 & 1.45\\
        E0: \autoref{fig:e2} & -1.62 & -1.75 & - & -\\
        E1: \autoref{fig:e2} & - & -2.94 & 0.74 & 1.18
    \end{tabular}
    \caption{Summary of results. A0 and B0 refers to 
    molecules belonging to \Tox{}, whereas C0, D0 and E0 belong to \Esol{}. Subsequent indexes refer to the related counterfactual explanations.}
    \label{tab:res}
\end{table}

\subsection{Explainability Results and Discussion}

We present some quantitative results in \autoref{tab:res}, listing some of the best counterfactual explanations collected for some samples in both tasks.
Both experiments have been tested with the convex combination between the neural encoding similarity and the Tanimoto, described in \eqref{eq:cc-sim}.

\begin{figure}
    \centering 
    \subfigure[A0]{
        \centering
        \includegraphics[width=.4\linewidth]{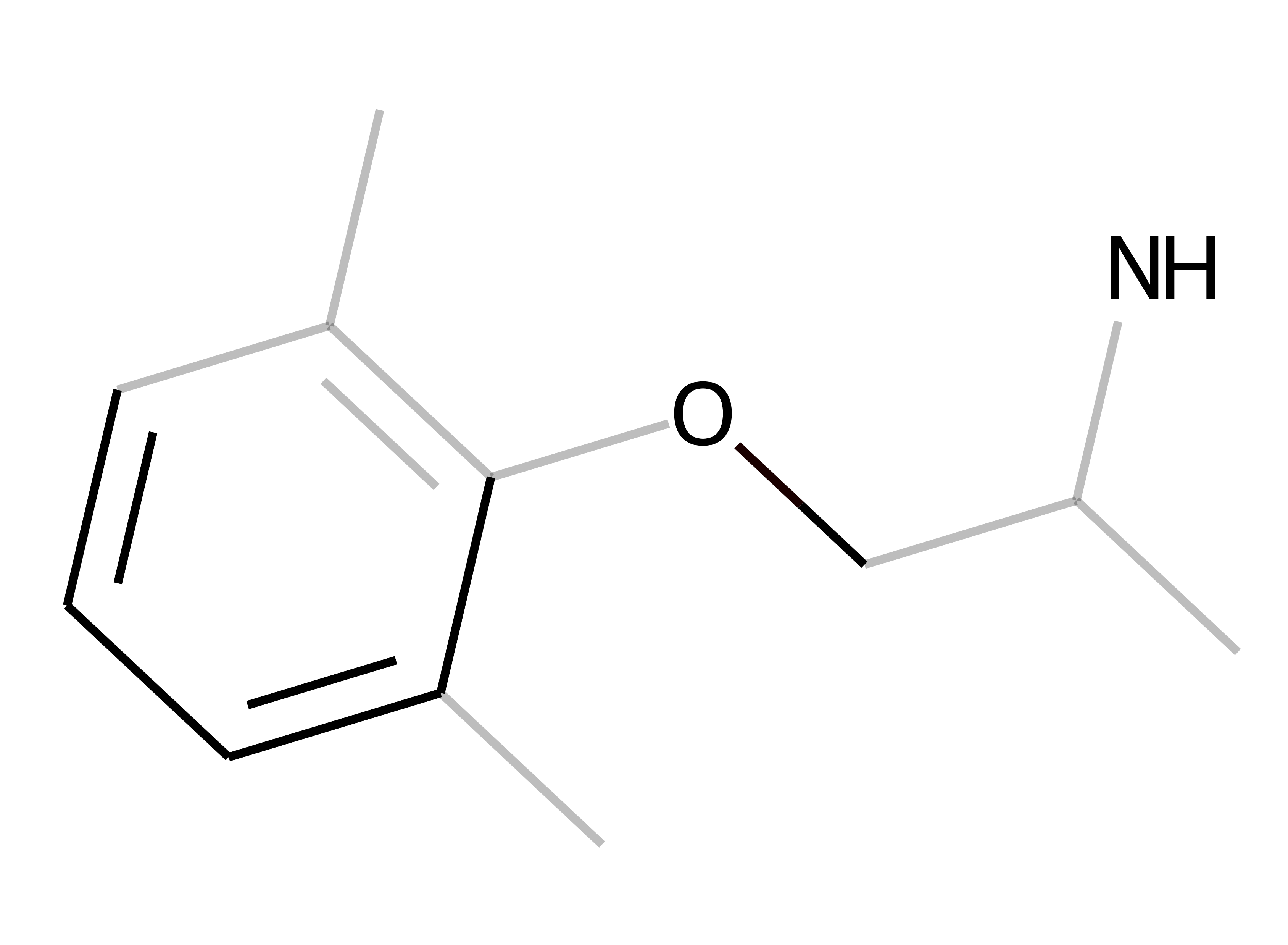}
    }
    \subfigure[A1]{
        \centering
        \includegraphics[width=.4\linewidth]{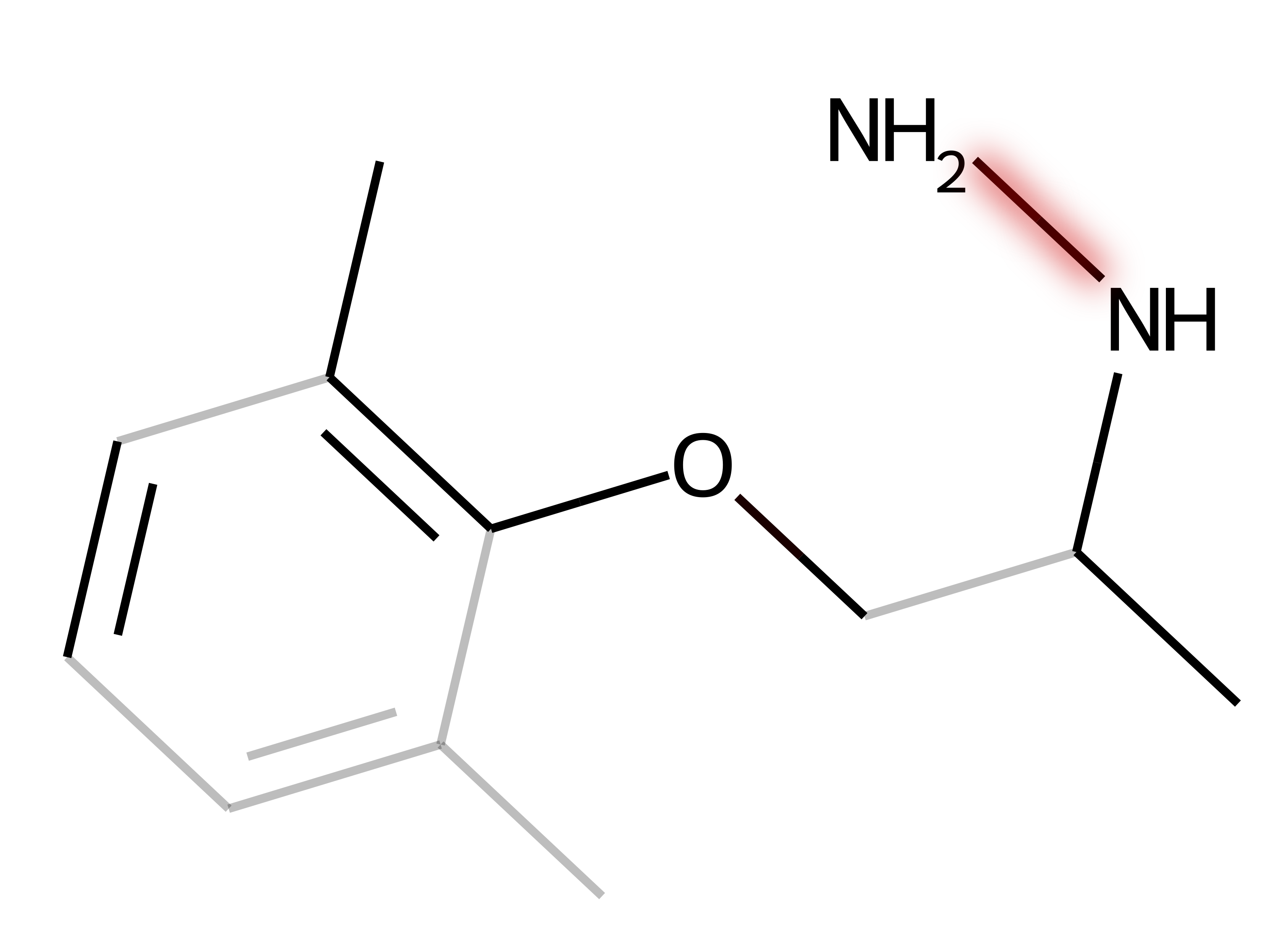}
    }
    \\[.1\baselineskip]
    \subfigure[A2]{
        \centering
        \includegraphics[width=.4\linewidth]{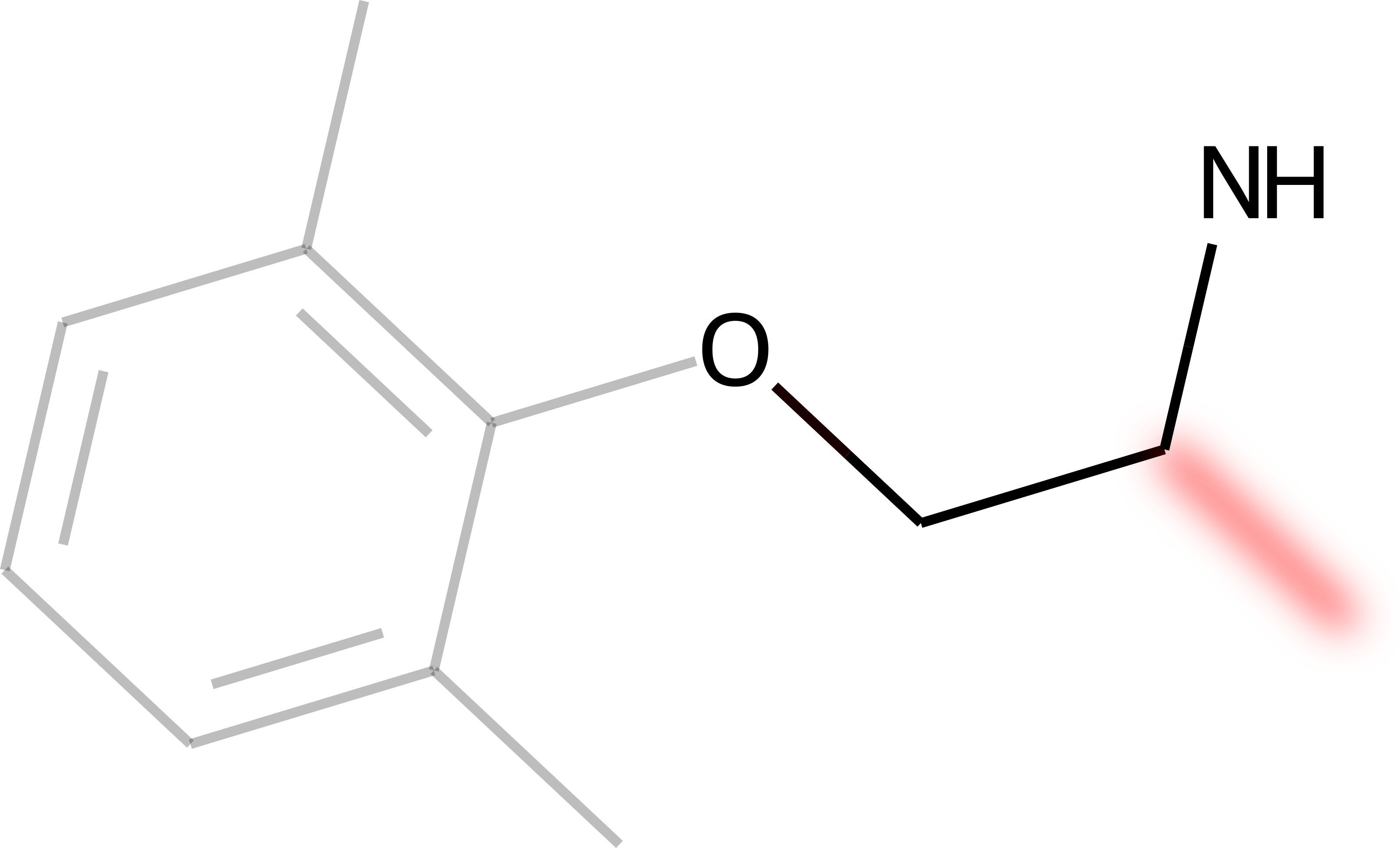}
    }
    \subfigure[A3]{
        \centering
        \includegraphics[width=.4\linewidth]{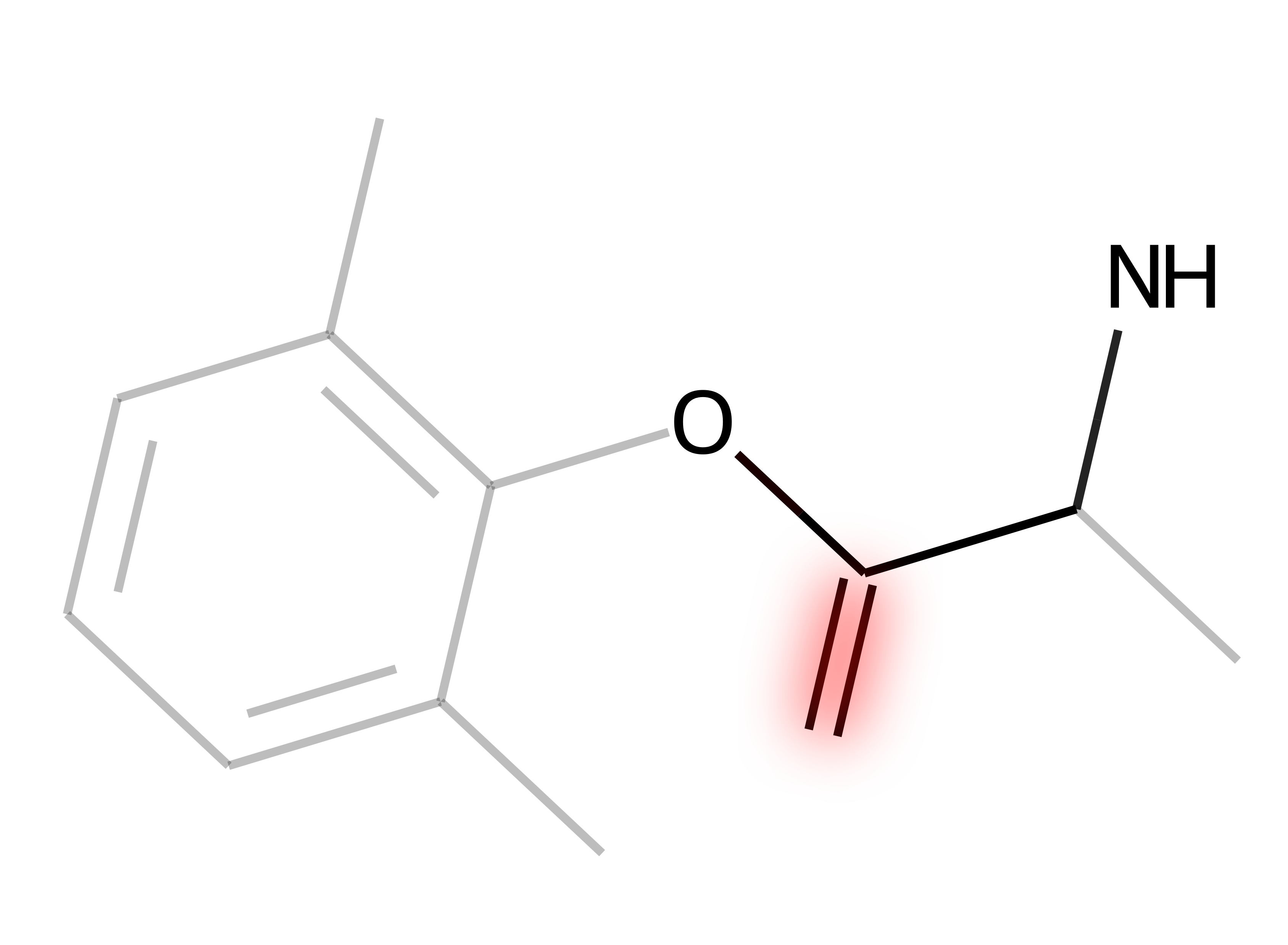}
    }
    \caption{Counterfactual explanations (A1-3) for a non-toxic molecule (A0).}
    \label{fig:t1}
\end{figure}
\begin{figure}
  \centering \subfigure[B0]{ \centering
  \includegraphics[width=.45\linewidth]{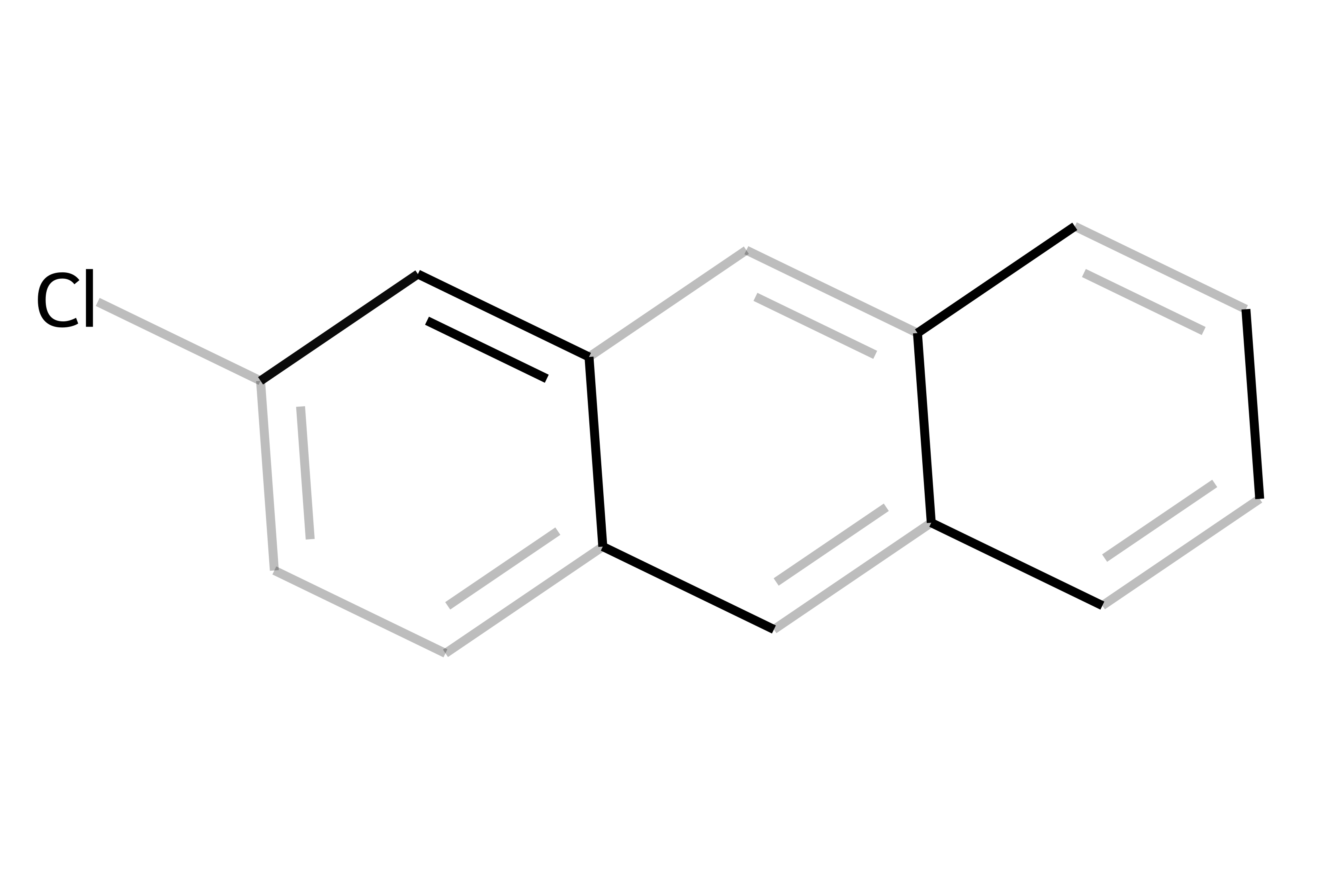} }
    \hspace{7pt} \subfigure[B1]{ \centering
      \includegraphics[width=.45\linewidth]{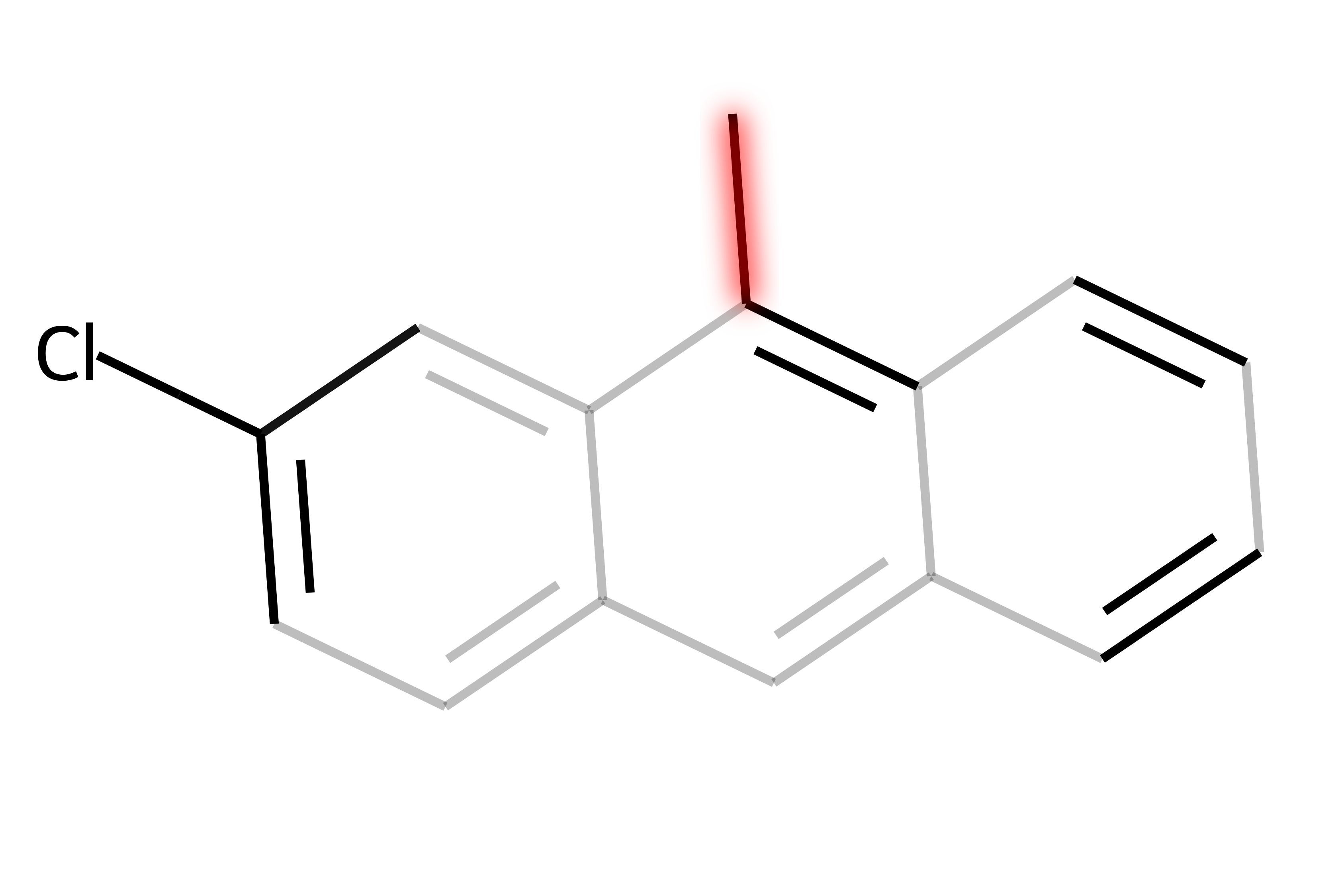}}
    \\[.1\baselineskip] \subfigure[B2]{ \centering
      \includegraphics[width=.45\linewidth]{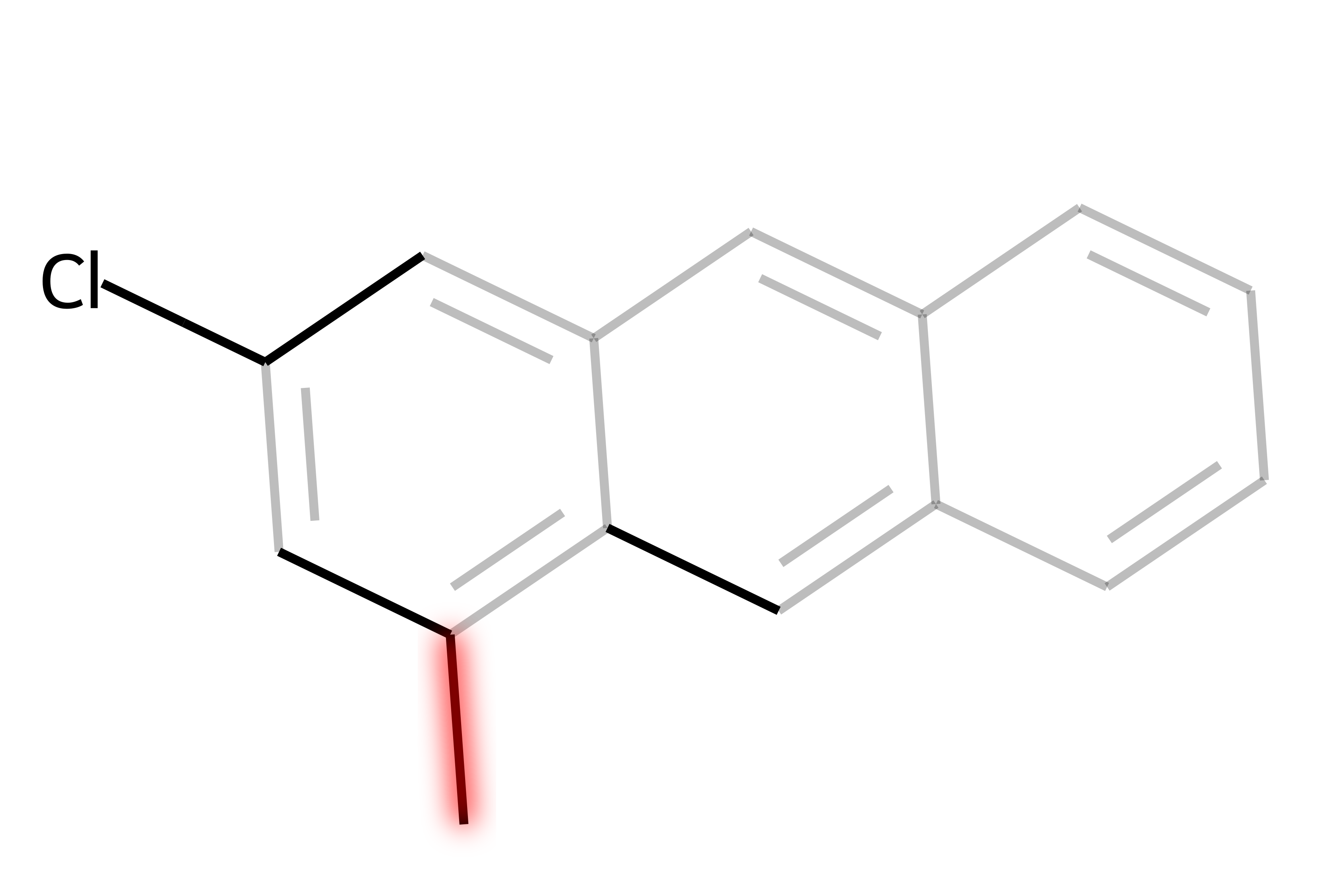}}
     \caption{Toxic molecule (B0) classified as non toxic after the addition of one atom of carbon (B1-2).}
    \label{fig:t2}
\end{figure}

\begin{figure}
  \centering \subfigure[C0]{ \centering
  \includegraphics[width=.4\linewidth]{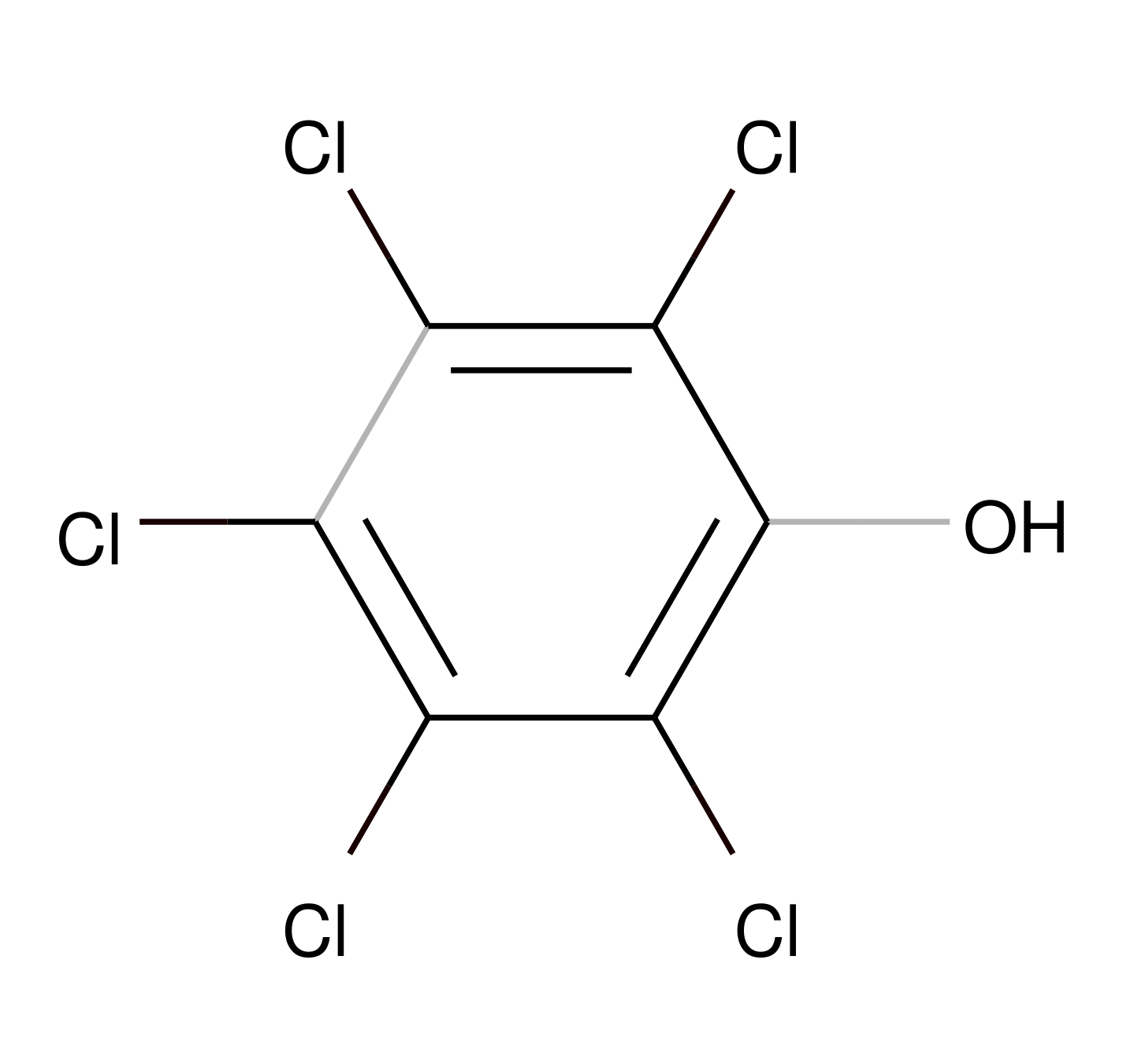} }
    \subfigure[C1]{ \centering
      \includegraphics[width=.4\linewidth]{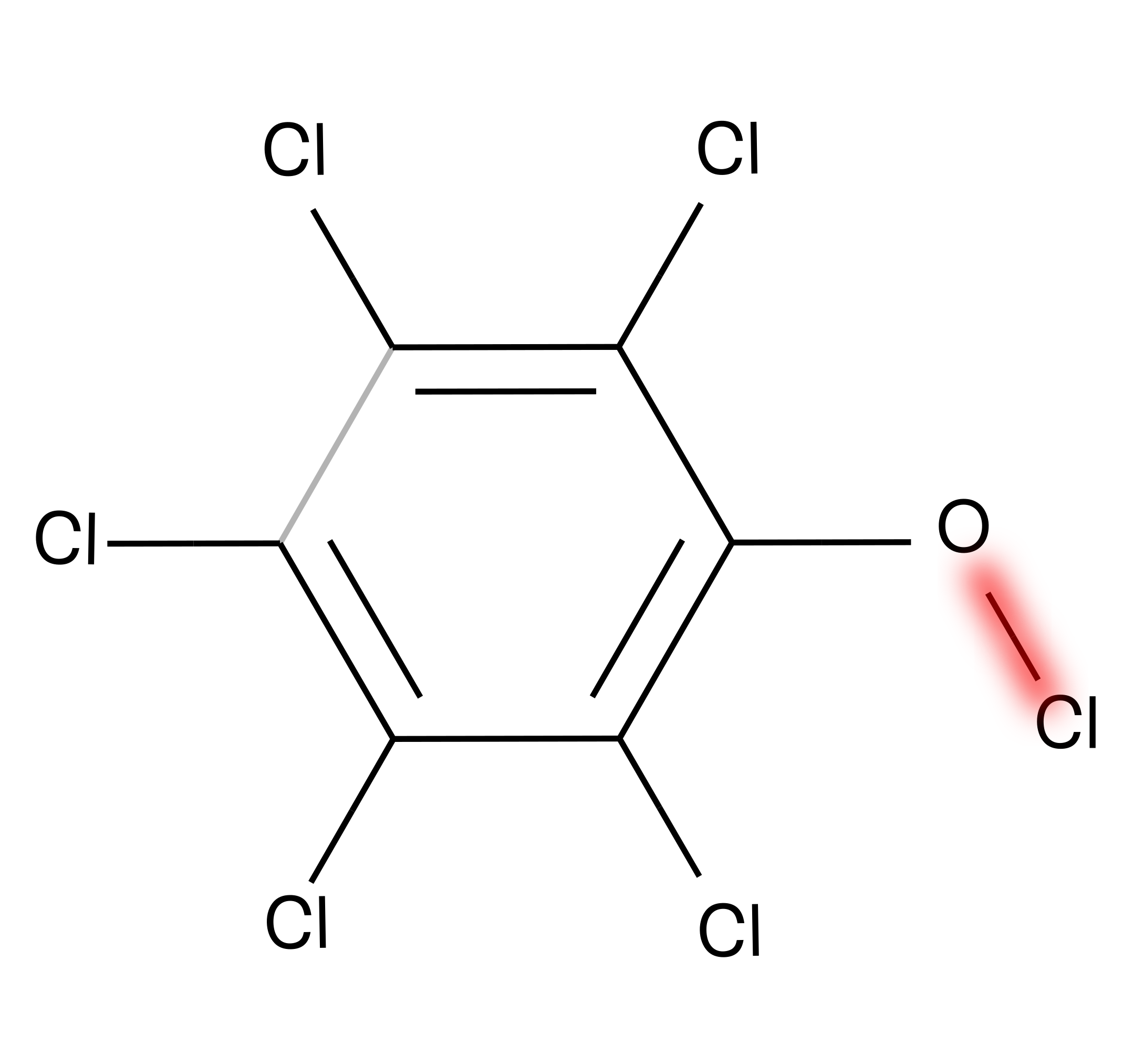} }
    \\[.1\baselineskip] \subfigure[C2]{ \centering
      \includegraphics[width=.4\linewidth]{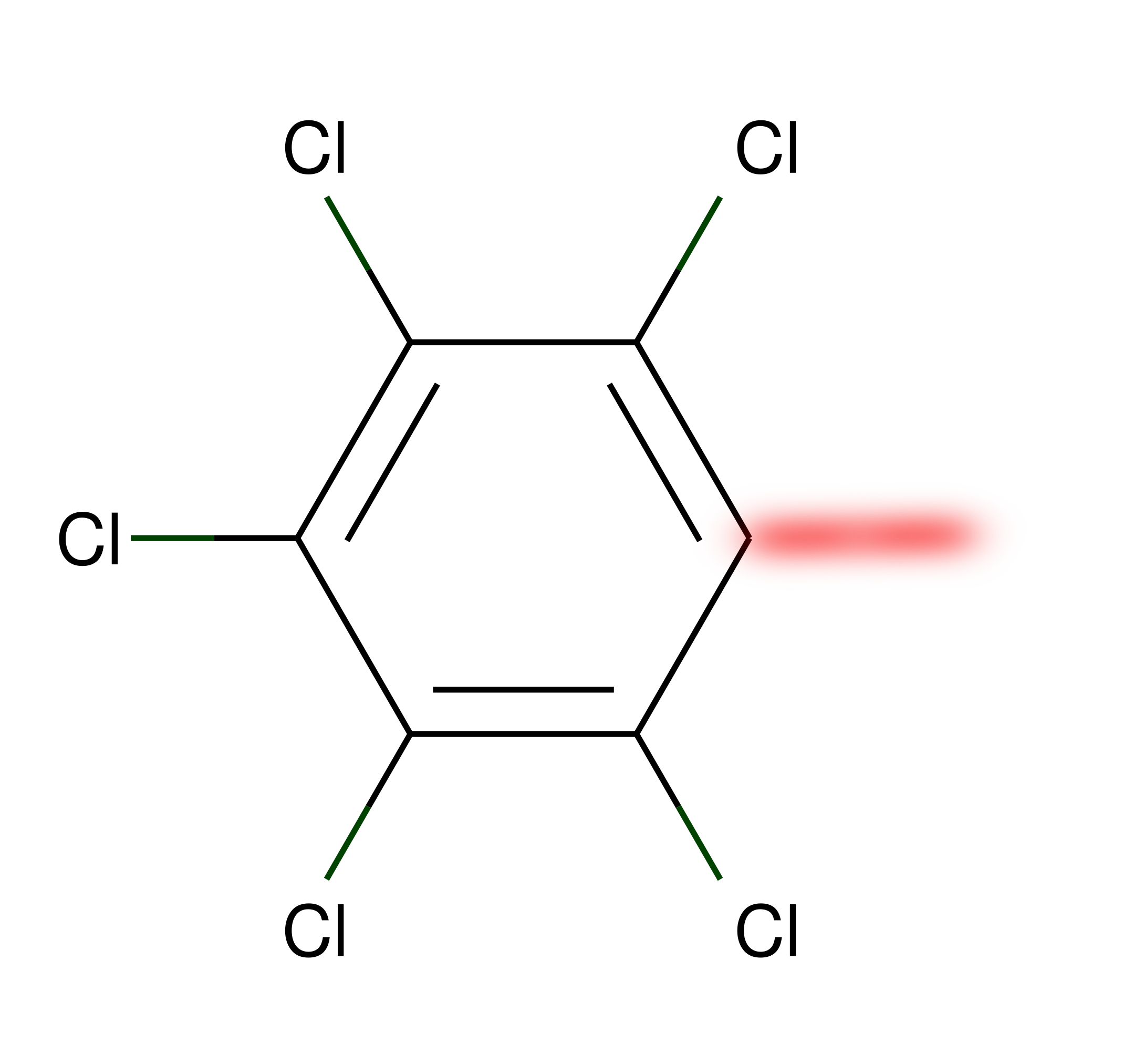} }
    \subfigure[C3]{ \centering
      \includegraphics[width=.4\linewidth]{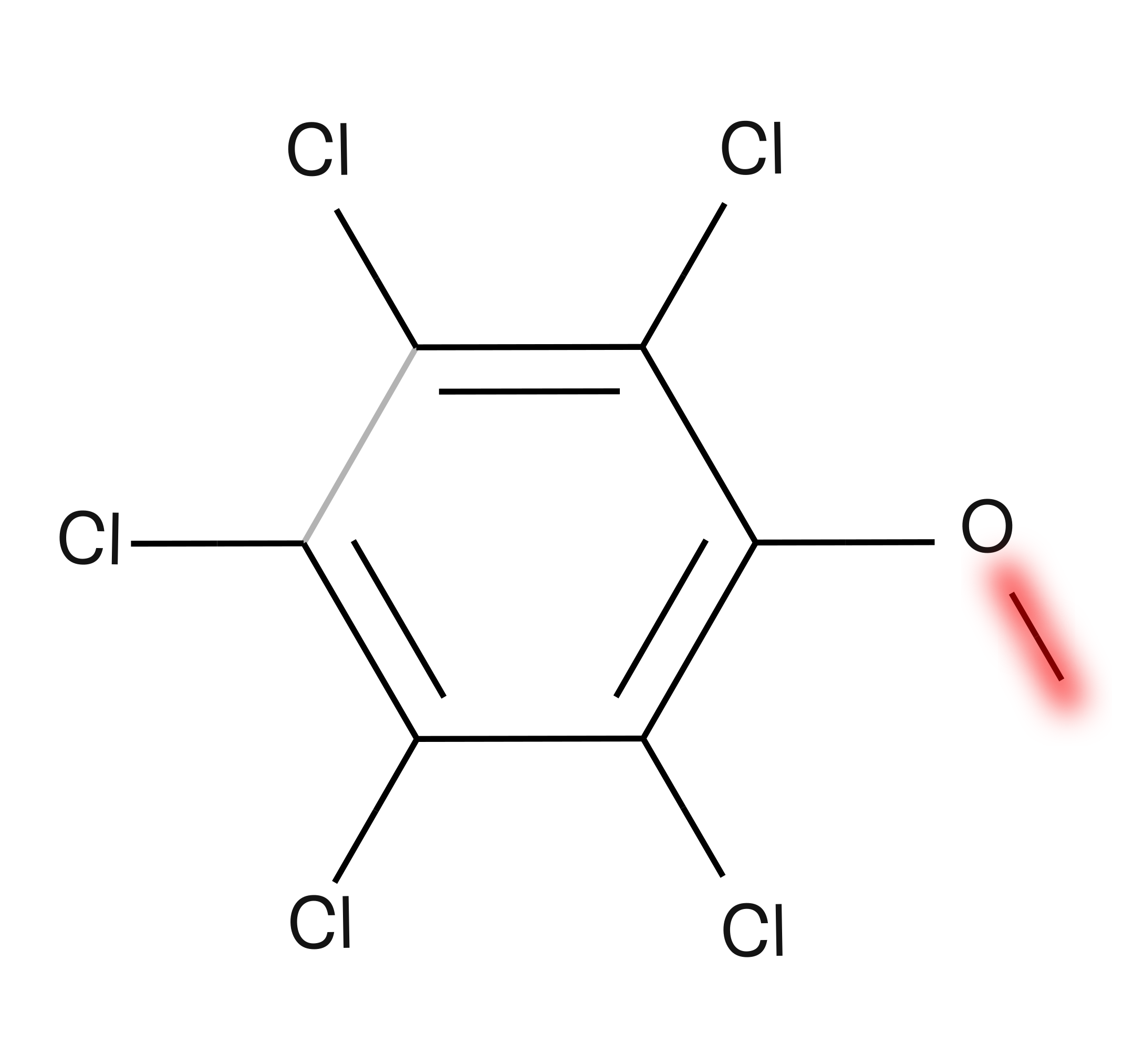} }
     \caption{ESOL sample alongside its counterfactuals
       (C1-3). Quantitative results are reported in
       \autoref{tab:res}.}
    \label{fig:e1}
\end{figure}

\begin{figure}
  \centering \subfigure[D0]{ \centering
  \includegraphics[width=.45\linewidth]{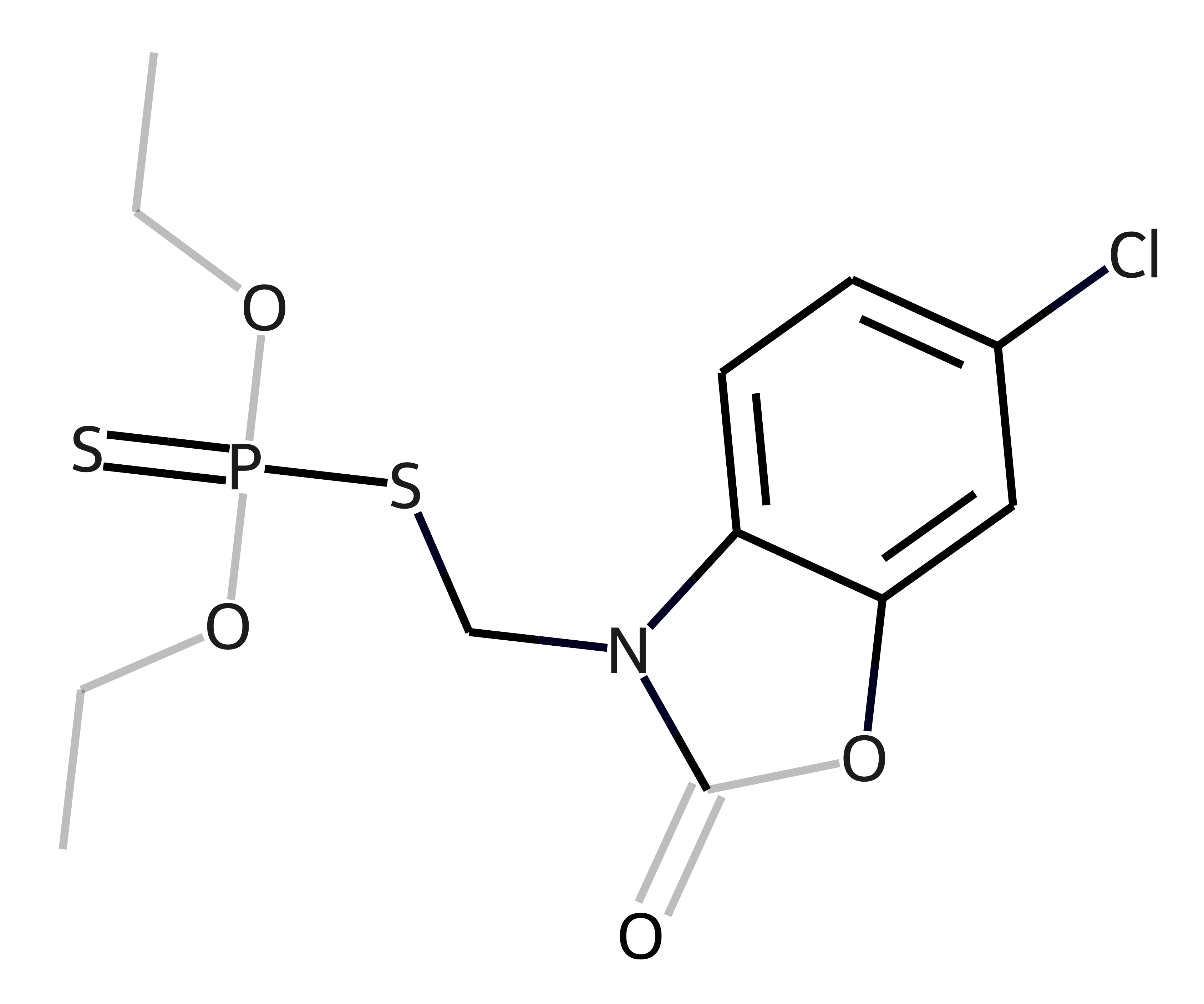} }
  \subfigure[D1]{ \centering
      \includegraphics[width=.45\linewidth]{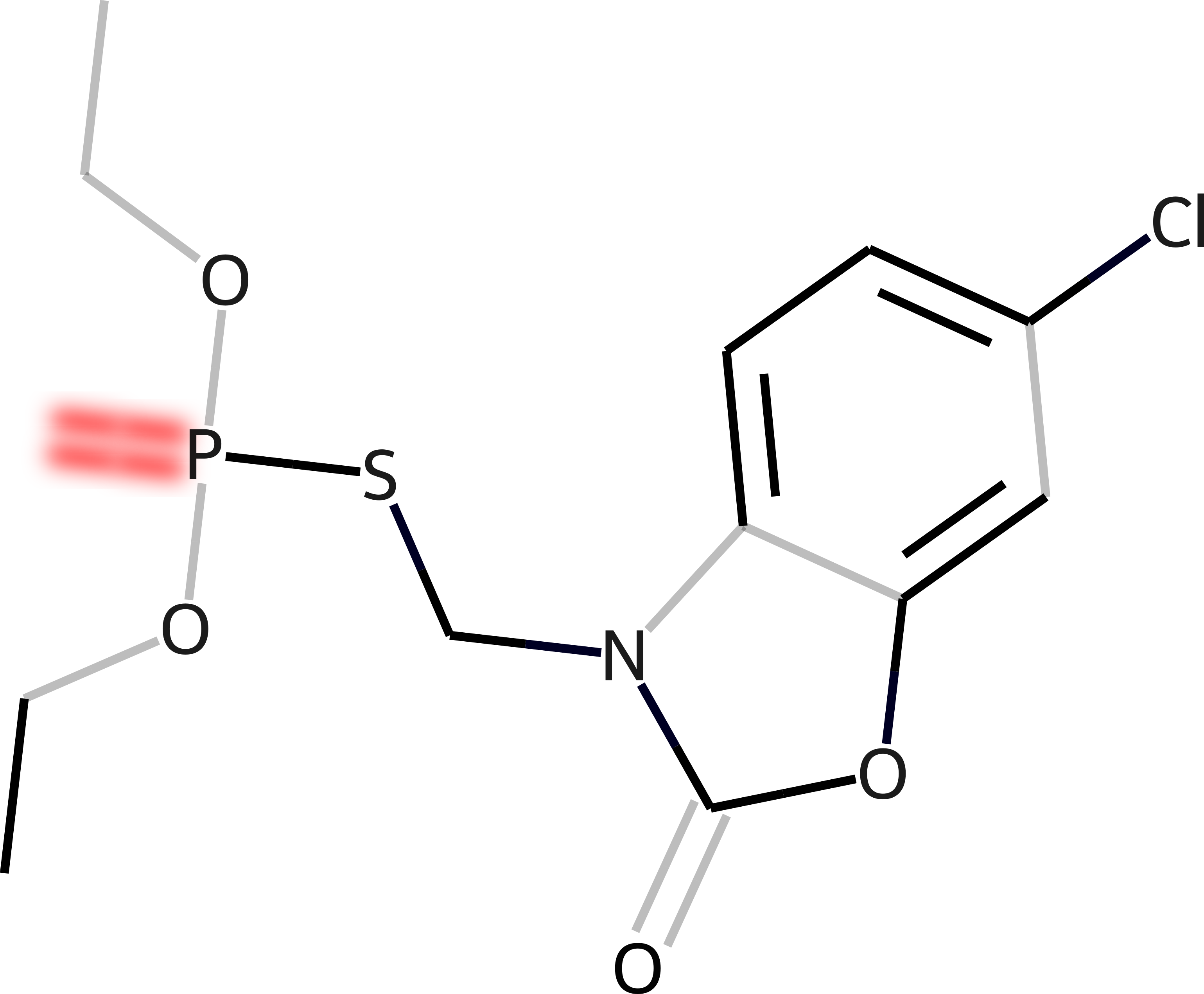}}
    \\[.1\baselineskip] \subfigure[E0]{ \centering
      \includegraphics[width=.45\linewidth]{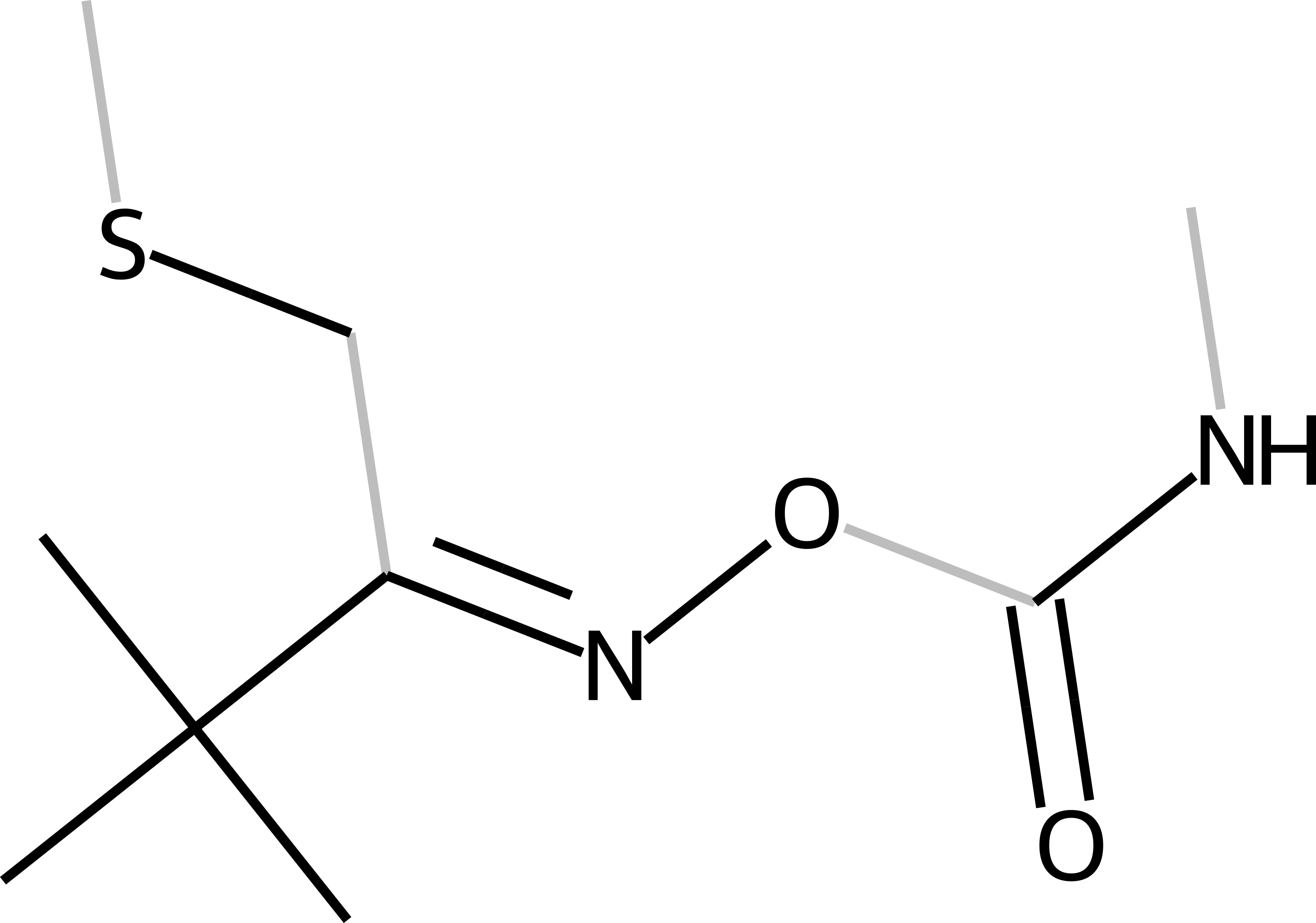}}
      \subfigure[E1]{ \centering
      \includegraphics[width=.4\linewidth]{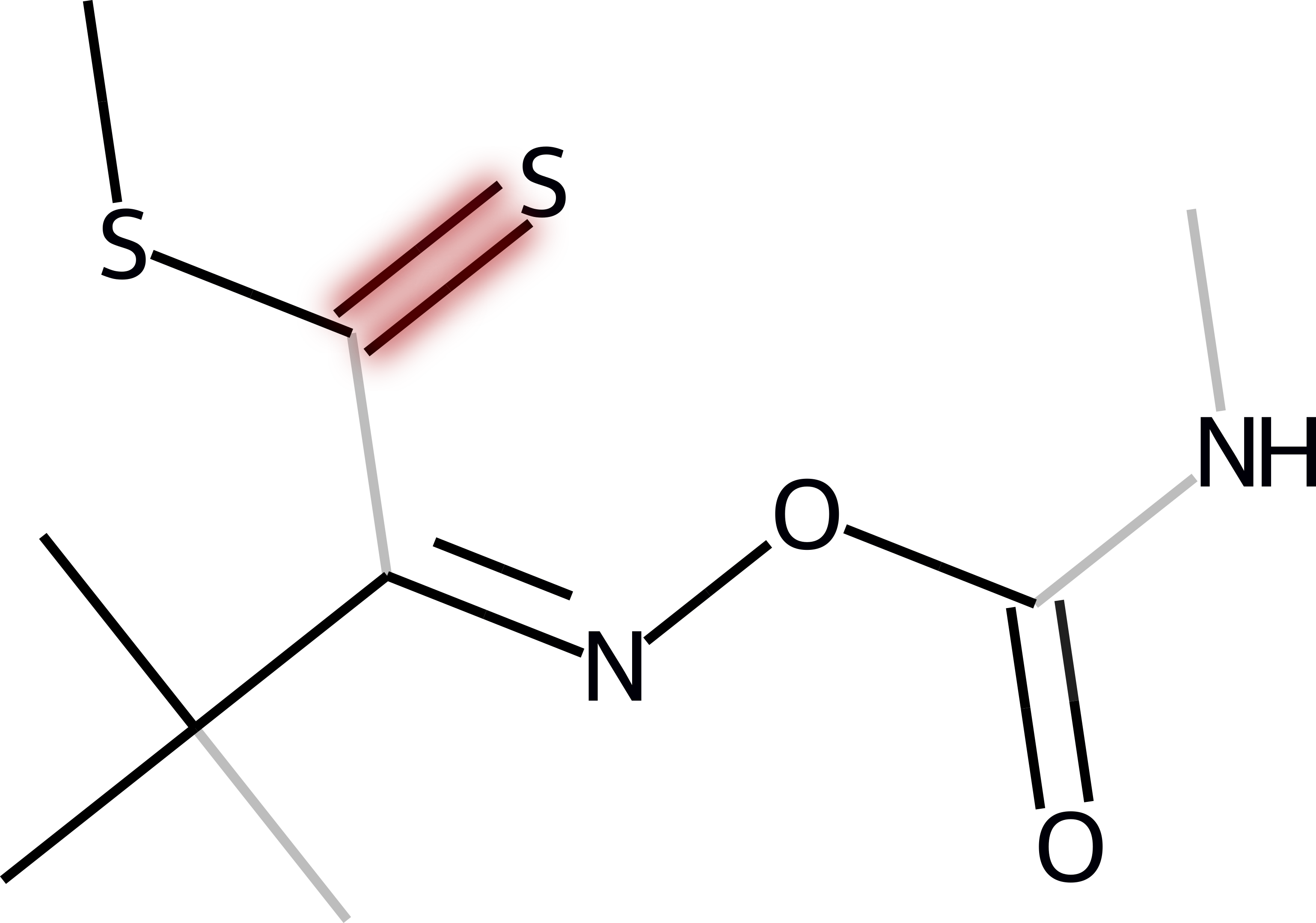} }
     \caption{First row: MEG removes sulphur, DGN solubility prediction decreases (\autoref{tab:res}).
     Second row: MEG adds sulphur, DGN predicts less solubility nonetheless.}
    \label{fig:e2}
\end{figure}

Qualitative results are shown in \autoref{fig:t1}, \autoref{fig:t2},
\autoref{fig:e1} and \autoref{fig:e2}. To ease the
interpretation of our results, counterfactual modifications have been
highlighted in red, while blurred edges represent those edges that
have been masked out by GNNExplainer predictions. In other words,
GNNExplainer interpretations are the sub-graphs formed by non-blurred
edges. 
By our examples, we:
\begin{itemize}
    \item highlight some of the qualitative differential reasoning that
    experts can carry out.
    \item show that GNNExplainer, by its very nature, generates explanations specifically
    tied with the input graph, making it difficult to find evidences in the global
    behaviour of the model. This is true for any model that follows closely the local explanation paradigm. We argue the necessity to output explanations that maintain their validity in a neighbourhood of the input graph of arbitrary size.
\end{itemize}

In \Tox{}, we analyse one representative result for each class of prediction.
In A0 (\autoref{fig:t1}), we take a molecule that has been correctly classified
by the DGN as being non-toxic and generate three counterfactuals for it.
Similarly, in B0 (\autoref{fig:t2}) we do the same for a toxic molecule. In both examples, the counterfactuals are classified as the opposite class with respect to the original one, as reported in \autoref{tab:res}.
In the first place, it is easy for domain experts to run qualitative analyses of the DGN predictions on our counterfactuals. In the first sample, we see
three valid modifications of A0, specifically the addition of an atom of nitrogen (A1), the removal of carbon (A2) and the addition of an atom of carbon linked to the molecule through a double bond (A3). In all these cases, it is sufficient to check whether these modifications can move the prediction from non-toxic to toxic also in a real-world scenario.
Correspondingly, in the toxic compound case (B0) the molecules resulting from addition of C, i.e B1-B2, are classified as non-toxic. 
Being able to highlight such behaviours is crucial when predicting safety-critical properties, e.g in drug design. 

Focusing on GNNExplainer interpretations, our counterfactuals show some difficulties that any local explainer based on single-instance explanation can encounter. First, we recall that GNNExplainer provides sample-level explanations by masking out low-importance edges in the graph. However, explanations limited to a single instance are difficult to generalise to other unexplained samples. This means that we have no clues on the out-of-instance model behaviour, making it 
difficult to guess how the model would react to unseen graphs. 
In fact, we see that in both \autoref{fig:t1}, \autoref{fig:t2} the original explanations
(A0-B0) have not identified as important certain substructures that have been eventually modified by our counterfactuals to change the prediction.

We now turn the attention to \Esol{} results, i.e C0-D0-E0 shown in
\autoref{tab:res}. C0 (\autoref{fig:e1}) is a well-known organic compound named pentachlorophenol,
commonly used as a pesticide or a disinfectant, and is characterised by nearly absolute insolubility in water. While the DGN achieved good predictive performance for its aqueous solubility value, the counterfactuals underlined that the models predicts less solubility in case the oxygen atom is removed (e.g, C2), or modified somehow (e.g, C1, C3), highlighting how it is highly relevant for the DGN prediction. As in the \Tox{} samples, such relation is not
adequately captured by GNNExplainer.

In the last sample, we bring a purely qualitative analysis of two molecules (\autoref{fig:e2}). In D0, we have a molecule of phosalone, while E0 is a molecule of thiofanox. MEG derives the two associated counterfactuals D1-E1 by performing complementary actions. In the first case, MEG finds D1 by removing an atom of sulphur from D0 while E1 is obtained by the addition of sulphur to E0. In both cases, though, the model prediction outputs lower values of solubility with respect to the original inputs (see \autoref{tab:res}). If the addition of sulphur makes the water solubility of a compound to decrease, the prediction of E1 is not faithful. Vice versa, if sulphur increases the water solubility, $\Model(D1)$ is not reliable.
Although $\Model(D0)$ and $\Model(E0)$ are precise with respect to the ground truth values, such an analysis shows indications that the learnt function may not be trustworthy.
It is our hope that, based on our interpretability approach, an expert of the molecular domain could be able to gain a better insight into the whether the properties and patterns captured by the predictive model are meaningful from a chemical standpoint.

\section{Conclusions}
We have presented MEG, a novel interpretability framework that tackles explainability in the chemical domain by generation of molecular counterfactual explanations. MEG can work with any DGN model as we only exploit input-output properties of such models. 

We show
As a general comment of the results, one can note that while a local approach such as GNNExplainer may give good approximations when it comes to explaining the specific prediction, it lacks sufficient breadth to characterize the model behaviour already in a near vicinity
of the sample under consideration. On the other hand, our
counterfactual interpretation approach may find new samples which are likely to highlight the causes of a given model prediction, providing a better approximation to a locally interpretable model, e.g. C1-3 in \autoref{fig:e1}.
Furthermore, MEG can help to detect some critical issues on toxicity prediction of a given model, e.g. B1-2 (\autoref{fig:t2}), that may affect humans' health if employed in the drug design setting.
In conclusion, apart for its value in generating
explanations that are well understood by a domain expert, MEG can work both as a sanity checker for other local model explainers, as well as a sampling method to strengthen the coverage and validity of local interpretable explanations, such as in the original LIME method for vectorial data \cite{lime}.

\section*{Acknowledgement}
The work is supported by project MIUR-SIR 2014 LIST-IT (grant n. RBSI14STDE) and by TAILOR, a project funded by EU Horizon 2020 research and innovation programme under GA No 952215.
\bibliographystyle{unsrt} 
\bibliography{ref} 

\begin{thebibliography}{10}

\bibitem{BacciuBLMOV19}
Davide Bacciu, Battista Biggio, Paulo Lisboa, Jos{\'{e}}~D. Mart{\'{\i}}n, Luca
  Oneto, and Alfredo Vellido.
\newblock Societal issues in machine learning: When learning from data is not
  enough.
\newblock In {\em 27th European Symposium on Artificial Neural Networks,
  {ESANN} 2019, Bruges, Belgium, April 24-26, 2019}, 2019.

\bibitem{micheli07}
Alessio Micheli, Alessandro Sperduti, and Antonina Starita.
\newblock An introduction to recursive neural networks and kernel methods for
  cheminformatics.
\newblock {\em Current Pharmaceutical Design}, 13(8), 2007.

\bibitem{zhou2018graph}
Jie Zhou, Ganqu Cui, Zhengyan Zhang, Cheng Yang, Zhiyuan Liu, Lifeng Wang,
  Changcheng Li, and Maosong Sun.
\newblock Graph neural networks: A review of methods and applications, 2018.

\bibitem{DBLP:journals/nn/BacciuEMP20}
Davide Bacciu, Federico Errica, Alessio Micheli, and Marco Podda.
\newblock A gentle introduction to deep learning for graphs.
\newblock {\em Neural Networks}, 129:203--221, 2020.

\bibitem{pmlr-v70-gilmer17a}
Justin Gilmer, Samuel~S. Schoenholz, Patrick~F. Riley, Oriol Vinyals, and
  George~E. Dahl.
\newblock Neural message passing for quantum chemistry.
\newblock volume~70 of {\em Proceedings of Machine Learning Research}, pages
  1263--1272, International Convention Centre, Sydney, Australia, 06--11 Aug
  2017. PMLR.

\bibitem{Errica2020A}
Federico Errica, Marco Podda, Davide Bacciu, and Alessio Micheli.
\newblock A fair comparison of graph neural networks for graph classification.
\newblock In {\em International Conference on Learning Representations}, 2020.

\bibitem{agrawal2015vqa}
A.~Agrawal, J.~Lu, S.~Antol, M.~Mitchell, C.~Lawrence Zitnick, D.~Batra, and
  D.~Parikh.
\newblock Vqa: Visual question answering, 2015.

\bibitem{zhao2020fast}
Yunxia Zhao.
\newblock Fast real-time counterfactual explanations, 2020.

\bibitem{Yuan2020}
Hao Yuan, Jiliang Tang, Xia Hu, and Shuiwang Ji.
\newblock Xgnn: Towards model-level explanations of graph neural networks.
\newblock {\em Proceedings of the 26th ACM SIGKDD International Conference on
  Knowledge Discovery \& Data Mining}, Jul 2020.

\bibitem{sutton1998reinforcement}
Richard~S Sutton and Andrew~G Barto.
\newblock Reinforcement learning: an introduction cambridge.
\newblock {\em MA: MIT Press.[Google Scholar]}, 1998.

\bibitem{micheli09}
A.~{Micheli}.
\newblock Neural network for graphs: A contextual constructive approach.
\newblock {\em IEEE Transactions on Neural Networks}, 20(3):498--511, 2009.

\bibitem{kipf2017semisupervised}
Thomas~N. Kipf and Max Welling.
\newblock Semi-supervised classification with graph convolutional networks,
  2017.

\bibitem{hamilton2017inductive}
William~L. Hamilton, Rex Ying, and Jure Leskovec.
\newblock Inductive representation learning on large graphs.
\newblock In {\em NIPS}, 2017.

\bibitem{bacciujmlr2020}
Davide Bacciu, Federico Errica, and Alessio Micheli.
\newblock Probabilistic learning on graphs via contextual architectures.
\newblock {\em Journal of Machine Learning Research}, 21(134):1--39, 2020.

\bibitem{simonyan2013deep}
Karen Simonyan, Andrea Vedaldi, and Andrew Zisserman.
\newblock Deep inside convolutional networks: Visualising image classification
  models and saliency maps, 2013.

\bibitem{zhou2015learning}
Bolei Zhou, Aditya Khosla, Agata Lapedriza, Aude Oliva, and Antonio Torralba.
\newblock Learning deep features for discriminative localization, 2015.

\bibitem{lime}
Marco~Tulio Ribeiro, Sameer Singh, and Carlos Guestrin.
\newblock "why should {I} trust you?": Explaining the predictions of any
  classifier.
\newblock In {\em Proceedings of the 22nd {ACM} {SIGKDD} International
  Conference on Knowledge Discovery and Data Mining, San Francisco, CA, USA,
  August 13-17, 2016}, pages 1135--1144, 2016.

\bibitem{cf}
Sandra Wachter, Brent Mittelstadt, and Chris Russell.
\newblock Counterfactual explanations without opening the black box: Automated
  decisions and the gdpr, 2018.

\bibitem{vanlooveren2020interpretable}
Arnaud~Van Looveren and Janis Klaise.
\newblock Interpretable counterfactual explanations guided by prototypes, 2020.

\bibitem{baldassarre2019explainability}
Federico Baldassarre and Hossein Azizpour.
\newblock Explainability techniques for graph convolutional networks.
\newblock {\em arXiv preprint arXiv:1905.13686}, 2019.

\bibitem{pope2019explainability}
Phillip~E Pope, Soheil Kolouri, Mohammad Rostami, Charles~E Martin, and Heiko
  Hoffmann.
\newblock Explainability methods for graph convolutional neural networks.
\newblock In {\em Proceedings of the IEEE Conference on Computer Vision and
  Pattern Recognition}, pages 10772--10781, 2019.

\bibitem{ying2019gnnexplainer}
Rex Ying, Dylan Bourgeois, Jiaxuan You, Marinka Zitnik, and Jure Leskovec.
\newblock Gnnexplainer: Generating explanations for graph neural networks,
  2019.

\bibitem{zhang2020relex}
Yue Zhang, David Defazio, and Arti Ramesh.
\newblock Relex: A model-agnostic relational model explainer, 2020.

\bibitem{huang2020graphlime}
Qiang Huang, Makoto Yamada, Yuan Tian, Dinesh Singh, Dawei Yin, and Yi~Chang.
\newblock Graphlime: Local interpretable model explanations for graph neural
  networks, 2020.

\bibitem{faber2020contrastive}
Lukas Faber, Amin~K. Moghaddam, and Roger Wattenhofer.
\newblock Contrastive graph neural network explanation, 2020.

\bibitem{nikolentzos2017matching}
Giannis Nikolentzos, Polykarpos Meladianos, and Michalis Vazirgiannis.
\newblock Matching node embeddings for graph similarity.
\newblock In {\em Proceedings of the AAAI Conference on Artificial
  Intelligence}, volume~31, 2017.

\bibitem{yuan2020interpretable}
Bo~Yuan, Ciyue Shen, Augustin Luna, Anil Korkut, Debora~S Marks, John Ingraham,
  and Chris Sander.
\newblock Interpretable machine learning for perturbation biology.
\newblock {\em bioRxiv}, page 746842, 2020.

\bibitem{azodi}
Christina~B. Azodi, Jiliang Tang, and Shin-Han Shiu.
\newblock Opening the black box: Interpretable machine learning for
  geneticists.
\newblock {\em Trends in Genetics}, 36(6):442 -- 455, 2020.

\bibitem{BACCIU2020177}
Davide Bacciu, Alessio Micheli, and Marco Podda.
\newblock Edge-based sequential graph generation with recurrent neural
  networks.
\newblock {\em Neurocomputing}, 416:177 -- 189, 2020.

\bibitem{PoddaBM20}
Marco Podda, Davide Bacciu, and Alessio Micheli.
\newblock A deep generative model for fragment-based molecule generation.
\newblock In Silvia Chiappa and Roberto Calandra, editors, {\em The 23rd
  International Conference on Artificial Intelligence and Statistics, {AISTATS}
  2020, 26-28 August 2020, Online [Palermo, Sicily, Italy]}, volume 108 of {\em
  Proceedings of Machine Learning Research}, pages 2240--2250. {PMLR}, 2020.

\bibitem{liu2015morl}
C.~{Liu}, X.~{Xu}, and D.~{Hu}.
\newblock Multiobjective reinforcement learning: A comprehensive overview.
\newblock {\em IEEE Transactions on Systems, Man, and Cybernetics: Systems},
  45(3):385--398, 2015.

\bibitem{sanchez2017}
Benjamin Sanchez-Lengeling, Carlos Outeiral, Gabriel~L. Guimaraes, and Alan
  Aspuru-Guzik.
\newblock Optimizing distributions over molecular space. an
  objective-reinforced generative adversarial network for inverse-design
  chemistry (organic), Aug 2017.

\bibitem{popova2018}
Mariya Popova, Olexandr Isayev, and Alexander Tropsha.
\newblock Deep reinforcement learning for de novo drug design.
\newblock {\em Science Advances}, 4(7):eaap7885, Jul 2018.

\bibitem{zhou2018optimization}
Zhenpeng Zhou, Steven Kearnes, Li~Li, Richard~N. Zare, and Patrick Riley.
\newblock Optimization of molecules via deep reinforcement learning, 2018.

\bibitem{rogers2010}
David Rogers and Mathew Hahn.
\newblock Extended-connectivity fingerprints.
\newblock {\em Journal of Chemical Information and Modeling}, 50(5):742--754,
  2010.
\newblock PMID: 20426451.

\bibitem{hasselt2015deep}
Hado van Hasselt, Arthur Guez, and David Silver.
\newblock Deep reinforcement learning with double q-learning, 2015.

\bibitem{dortmund2016}
Kristian Kersting, Nils~M. Kriege, Christopher Morris, Petra Mutzel, and Marion
  Neumann.
\newblock Benchmark data sets for graph kernels, 2016.

\bibitem{wu2017moleculenet}
Zhenqin Wu, Bharath Ramsundar, Evan~N. Feinberg, Joseph Gomes, Caleb Geniesse,
  Aneesh~S. Pappu, Karl Leswing, and Vijay Pande.
\newblock Moleculenet: A benchmark for molecular machine learning, 2017.

\bibitem{landrum2006rdkit}
Greg Landrum et~al.
\newblock Rdkit: Open-source cheminformatics, 2006.

\bibitem{morris2018weisfeiler}
Christopher Morris, Martin Ritzert, Matthias Fey, William~L. Hamilton, Jan~Eric
  Lenssen, Gaurav Rattan, and Martin Grohe.
\newblock Weisfeiler and leman go neural: Higher-order graph neural networks,
  2018.

\bibitem{Fey/Lenssen/2019}
Matthias Fey and Jan~E. Lenssen.
\newblock Fast graph representation learning with {PyTorch Geometric}.
\newblock In {\em ICLR Workshop on Representation Learning on Graphs and
  Manifolds}, 2019.

\end{thebibliography}
\include{appendix}

\end{document}